\documentclass[letterpaper]{article} %
\usepackage[preprint]{aaai2027}
\usepackage[hyphens]{url}  %
\usepackage{graphicx} %
\usepackage{natbib}  %
\usepackage{caption} %
\usepackage{algorithm}
\usepackage{algorithmic}

\usepackage{newfloat}
\usepackage{listings}
\DeclareCaptionStyle{ruled}{labelfont=normalfont,labelsep=colon,strut=off} %
\floatstyle{ruled}
\newfloat{listing}{tb}{lst}{}
\floatname{listing}{Listing}

\usepackage{booktabs}

\usepackage{amsmath}
\usepackage{amssymb}
\usepackage{multirow}
\usepackage[capitalize]{cleveref} 
\usepackage{enumitem}
\usepackage{booktabs}

\title{What Factors Matter for Collection-Level Counting with Human--AI Collaboration?}
\author {
    Seoyun Jeong\textsuperscript{\rm 1},
    Logan Lawrence\textsuperscript{\rm 1},
    Owen L. Martin\textsuperscript{\rm 2},
    Benjamin Koger\textsuperscript{\rm 2},
    Daniel Sheldon\textsuperscript{\rm 1},
    Subhransu Maji\textsuperscript{\rm 1}
}
\affiliations {
    \textsuperscript{\rm 1}University of Massachusetts Amherst\\
    \textsuperscript{\rm 2}University of Wyoming\\
    seoyunjeong@umass.edu, lclawrence@umass.edu, omartin9@uwyo.edu, bkoger@uwyo.edu, sheldon@cs.umass.edu, smaji@umass.edu
}

\begin{document}

\maketitle

\begin{abstract}
Many scientific and policy decisions require estimating the total number of target objects across a large image collection. Exhaustive human labeling is often infeasible, while raw model predictions can be biased, especially under domain shift. 
Recent human--AI estimation methods combine imperfect model predictions with limited human verification to obtain reliable collection-level estimates. 
However, existing work provides little guidance on how practitioners should choose among models, estimators, and adaptation strategies for a new deployment. 
We study these choices within a common human-in-the-loop counting framework on CLC-12, a benchmark of 12 tasks spanning biomedical microscopy, ground-level and aerial wildlife surveys, and overhead imagery of urban structures.
Our results show that modern multimodal models provide a strong, training-free starting point for many tasks, while fine-tuned detectors become more advantageous at larger labeling budgets and across repeated deployments.
Estimation efficiency improves further when model predictions are used to prioritize which units humans label, rather than sampling uniformly.
We also find that each model's ranking quality can be estimated with a small pilot set, yielding practical guidance for selecting strategies on new counting tasks.
In a field deployment on aerial pronghorn surveys, a task-specific adaptation of these methods meets the survey's $\pm 5\%$ tolerance after reviewing a median of only $3.2\%$ of collected images.
\end{abstract}

\begin{links}
    \link{Code}{https://github.com/cvl-umass/collection-counting-benchmark}
\end{links}

\section{Introduction}

\begin{figure}[t]
    \centering
    \includegraphics[width=0.9\columnwidth]{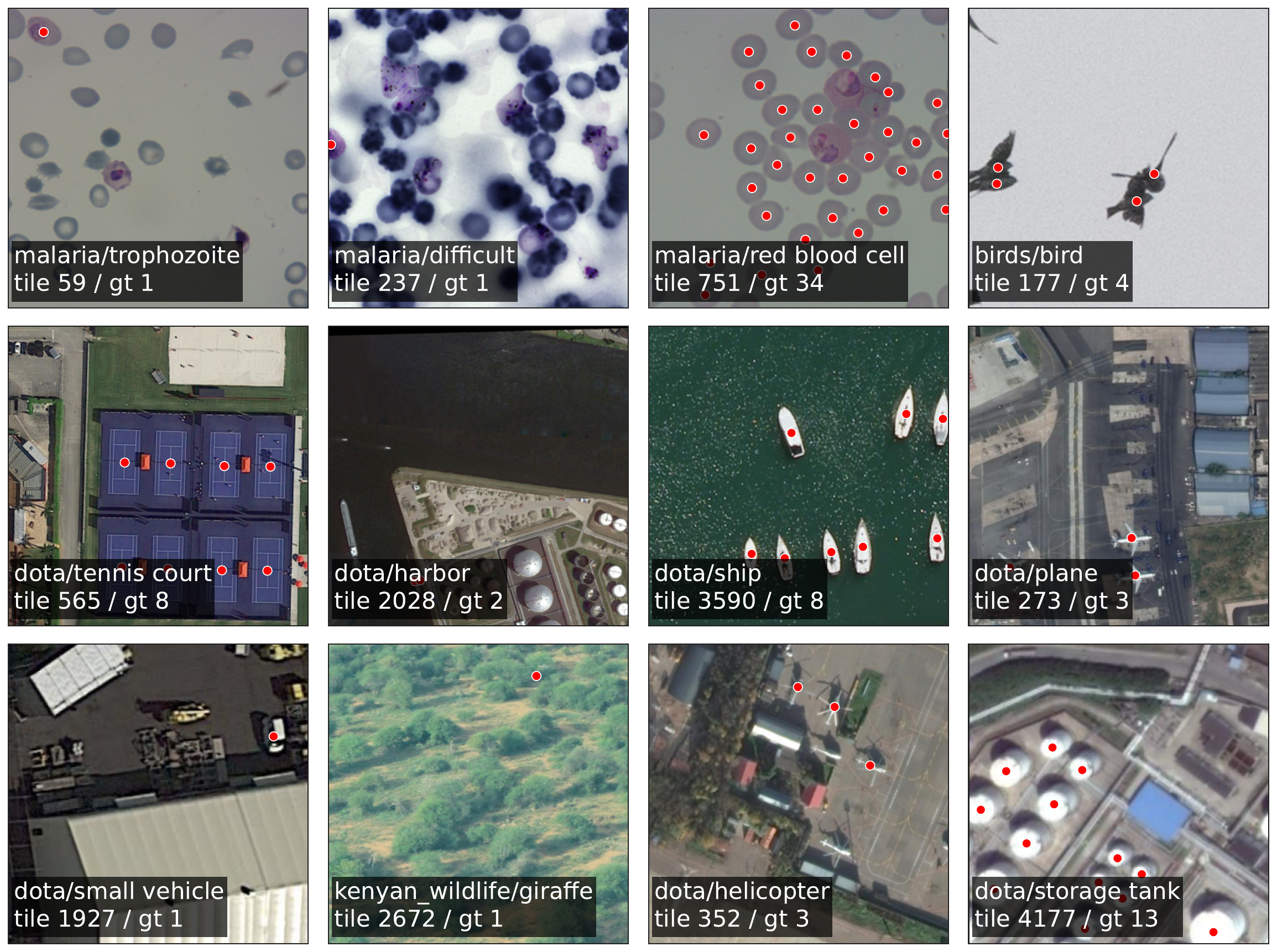}
    \caption{Samples from the 12 counting tasks. Each in-image label identifies the task (dataset/class), tile index, and exact ground-truth count. Red points mark each object.}
    \label{fig:tile_montage}
\end{figure}

Many scientific and policy questions involve counting in image collections: How many infected cells are present in a set of blood smears? How many birds migrate through a region in a given year? How many buildings were damaged in a disaster-affected region? Such totals can inform public-health surveillance~\citep{rajaraman2018malaria}, conservation policy~\citep{norouzzadeh2018wildanimals, ds_kenyan_wildlife}, and disaster response~\citep{gupta2019xbd}, aiding allocation of limited resources.
However, in each case, the imagery needed to answer these questions is abundant, while exhaustive human review does not scale to the collection sizes now routinely produced by their imaging pipelines. 
Similar problems arise in many other social impact domains where practitioners often have limited resources for data labeling and model development but require high precision or statistical guarantees that are not easily achieved by pure AI methods.

A growing body of work combines imperfect model predictions with a small amount of human-verified labels to obtain statistically valid estimates of population totals \citep{meng2022iscount, angelopoulos2023predictionpoweredinference, perez2023discount, hamilton2025activemeasurement}. 
The high-level idea is to use the model to make predictions on the entire collection, then sample images for human labeling to correct the model's total within a statistical estimation framework. 
These methods can provide unbiased estimates and confidence intervals regardless of the predictor's accuracy. However, their \emph{efficiency}---how much human labeling is needed to achieve a desired level of precision---depends strongly on the predictor model, estimator, data distribution, and labeling protocol.

Prior work offers limited practical guidance on selecting human--AI counting strategies under different deployment constraints. Practitioners may begin with no training data, a detector trained on a related domain, or sufficient resources to train and adapt a custom model. These settings create tradeoffs among model quality, labeling efficiency, computational cost, and engineering effort. 
We therefore ask: Given a target image collection and a set of resource constraints, which model and estimator should one use, and when is adaptation worthwhile? 
Alongside a new benchmark, we demonstrate the practical value of this model-assisted counting framework on real-world aerial pronghorn surveys, where counting from imagery rather than physically from the aircraft removes the most dangerous task wildlife professionals perform \cite{sasse2003job}.
Our contributions are:

\begin{itemize}[leftmargin=*,topsep=1pt,itemsep=0pt]
\item \textbf{CLC-12}, a collection-level counting benchmark with 12 tasks across biomedical microscopy, wildlife survey imagery, and overhead urban imagery (\cref{fig:tile_montage}).
\item A controlled comparison of models, estimators, and adaptation strategies under a common human-in-the-loop estimation framework, together with a procedure for pilot-based model selection.
\item A case study on aerial pronghorn surveys that puts these findings into practice, including a lightweight adaptation scheme for repeated deployments.
\end{itemize}
In preview, we find that modern off-the-shelf open-source multimodal large language models are a remarkably strong starting point for model-assisted counting.
On nearly every task in our benchmark, they reach counting error below 10\% with under 10\% of the collection labeled, cutting error by roughly 4$\times$ relative to a model-free estimate that uses only the human-labeled units, at the same budget. 
Training a custom detector can lower the error further, but at the cost of larger labeling budgets and model training time.
On the pronghorn survey, $\pm5\%$ uncertainty tolerance is achieved after reviewing a median of only 3.2\% of the collected images.

\section{Related Work}

Counting objects in images has been studied for well over a decade. Detection-based methods localize and then add up each instance using general object detectors such as Faster R-CNN~\cite{ren2015fasterrcnn}, while density-based methods instead regress a per-pixel density whose integral gives the count, sidestepping explicit detection in crowded scenes~\cite{lempitsky2010countobjects, zhang2016multicolumncrowd}. 
More recently, class-agnostic models extend to novel categories from a few exemplars \cite{ranjan2021learningcount} or from text queries, as in open-vocabulary detectors (OVDs) and multimodal large language models (MLLMs), with a growing set of benchmarks measuring their reliability \cite{paiss2023teachingclipcount, li2025countsbench}.
Across these paradigms, evaluation targets accuracy on individual images, rather than a total across an entire collection.
We instead benchmark models by how well they support estimating collection-level totals, a criterion that does not always track per-image counting accuracy, as our experiments show.

A complementary line of work pairs imperfect model predictions with a small number of human-verified labels to obtain statistically valid estimates of a collection-level total. Its foundations lie in survey sampling, where auxiliary variables correlated with the target yield ratio, regression, and other model-assisted estimators~\cite{cochran1977sampling, sarndal2003model}. Recent methods replace hand-chosen covariates with learned predictors: IS-Count~\cite{meng2022iscount} and DISCount~\cite{perez2023discount} sample images by importance weights derived from model scores, and active measurement~\cite{hamilton2025activemeasurement} interleaves labeling, estimation, and model adaptation in a sequential loop. Prediction-powered inference~\cite{angelopoulos2023predictionpoweredinference} extends the same principle to general estimands, correcting black-box predictions with a labeled subset to form valid confidence intervals, and the related active-testing literature draws informative samples to evaluate models label-efficiently~\cite{kossen2022active, farquhar2021statistical}. 
Our work builds on advances in both counting models and statistical estimators. 
We evaluate both fine-tuned and frozen models, and consider practical factors such as inference costs and required software expertise to adapt models to target distributions.

\section{Estimating Collection-Level Counts}

\subsection{Problem Setup}\label{sec:setup}

We study estimating a single-class \emph{collection-level total}, i.e., the number of target objects across an image collection. The collection is represented as a finite pool $\Omega$ of $N=|\Omega|$ units. Each unit $s\in\Omega$ is a cropped image tile or, for images too small to tile, the whole image (\S\ref{sec:tasks}). Let $f(s)\in\mathbb{Z}_{\ge 0}$ denote the ground-truth count in unit $s$ and $F(S) = \sum_{s \in S} f(s)$ be the total of any subset $S \subseteq \Omega$. 
The target is
$
F_\text{true} \equiv F(\Omega)=\sum_{s\in\Omega} f(s).
$
We formulate this problem within \citet{hamilton2025activemeasurement}'s active measurement framework, in which true per-unit counts are acquired sequentially and combined with model predictions to estimate a collection-level total. It generalizes DISCount~\cite{perez2023discount} to allow model adaptation and sampling without replacement. These features translate well to deployment settings where practitioners accumulate labels and never label the same unit twice.

Assume that a model provides a predicted count $g(s) \ge 0$ for each unit. 
Before the main labeling loop, fine-tuned detectors are conditioned with a labeled warmup set $\mathcal W$, where at least half contain one or more instances of the target object. For frozen models, $\mathcal W = \emptyset$.
At time steps $t=1,2,\ldots, T$, let
$
  \mathcal D_t = \mathcal W \cup \{s_1,\ldots,s_{t-1}\}
$
denote the units revealed before the step. A sampler draws an unlabeled unit
$
  s_t \sim q_t
$
from the remaining pool $\mathcal R_t = \Omega\setminus \mathcal D_t$, where $q_t$ is the current proposal distribution over the remaining pool. 
Then $f(s_t)$ is revealed, corresponding to a human annotation. 
Next, an estimator combines the true counts on $\mathcal D_t^+ = \mathcal D_t\cup\{s_t\}$ with the model predictions $g_t$ available at step $t$, where $g_t(s)$ is the predicted count on unit $s$, to produce an estimate $\hat F$ of the total. For frozen models, $g_t = g_0$ for all $t$.
Finally, $\mathcal D_t^+$ is optionally used to fine-tune the model.
To summarize, a counting strategy consists of (i) a prediction model that produces $g_t$, (ii) a proposal distribution (i.e., sampler) $q_t$, (iii) an estimator for the total, and (iv) a model-update policy. 

\subsection{Models}\label{sec:models}
The framework requires only a model that produces a per-unit count; any such model can be substituted. We consider two groups of models a practitioner might choose for a new task: detectors fine-tuned on the target data, and larger frozen models used off the shelf.

\subsubsection{Fine-tuned detectors} 
We consider two Faster R-CNN detectors~\cite{ren2015fasterrcnn}, one with a larger ResNet-50~\cite{he2016resnet50} backbone and one with a smaller MobileNetV3-Large~\cite{howard2019mobilenet} backbone. Both are initialized from COCO-pretrained weights and adapted to a single foreground class by replacing the box-prediction head with a binary object-versus-background head. 
They are initially fine-tuned on the warmup set of $w=50$ units, and continually retrained every $25$ steps during the sequential labeling loop. Training details are provided in Appendix~\ref{appendix:frcnn_training}.

\subsubsection{Frozen OVDs and MLLMs}
Given an input image and a target object name, OVDs return localized detections. We use the number of detections above a 0.5 confidence threshold as the per-unit count. We evaluate three OVDs: Florence-2 \cite{xiao2023florence}, Grounding DINO \cite{liu2023groundingdino}, and OWLv2 \cite{minderer2023owlv2}.

MLLMs take an image and a natural-language prompt and respond in free-form text. 
For all MLLMs, we use the prompt \emph{``How many \{object\_name\} are in this image?''} and extract the count using a constrained LLM-based answer extractor \cite{Lawrence_2026_WACV}. Predictions without a clear numeric answer are assigned a count of zero. 
To compare locally runnable open-source models with proprietary models, we evaluate five open-source models (Qwen3-VL-8B \cite{bai2025qwen3}, InternVL3.5-8B \cite{wang2025internvl3}, Molmo-7B-D-0924 \cite{deitke2025molmo}, Molmo2-8B \cite{clark2026molmo2}, Aya Vision-8B \cite{dash2025aya}) and three commercial models (GPT-5.2 \cite{openai2025gpt52}, Gemini 3.1 Pro \cite{google2026gemini31pro}, Qwen3.7-Plus \cite{qwen2026qwen37plus}).
Model-specific inference and cost details are in Appendix~\ref{appendix:mllm_prompting_details}.

\subsection{Estimators and Sampling Distribution}\label{sec:estimators}
We compare four ways of using the model predictions $g_t$: trust them on the unlabeled pool, ignore them, correct them with uniformly sampled units, or use them to direct label acquisition. All sampling is without replacement. Except for the model-only baseline, the estimators are unbiased (see Appendix~\ref{appendix:estimator_math}).
Our estimators use two different strategies: 
\emph{batch} estimators use the entire sampled set $\mathcal D_t^+$ in each step to form an estimate $\hat F_{1:t}$; \emph{incremental} estimators form separate estimates $\hat F_t$ in each step and later combine per-step estimates into an overall estimate. 
Below, we derive expressions for the total; analogous expressions for the variance can be used to construct confidence intervals (\citealt{mcbook}; Appendix~\ref{appendix:estimator_math}).

\subsubsection{Model-only baseline}

\emph{Can we trust the model alone?}
For a fair sequential comparison, the model-only baseline replaces predictions on revealed units with their exact counts and trusts the active model on the remainder. Writing $G_t(S)=\sum_{s\in S}g_t(s)$, the batch estimator is:
\begin{equation}
    \hat F_{1:t}^{\mathrm{model}} = F(\mathcal D_t^+)+ G_t(\Omega\setminus\mathcal D_t^+).
\label{eq:model_only}
\end{equation}
\citet{kossen2022active} adopt the same approach, which they call model-based \emph{interpolation}, for active testing.

\subsubsection{Model-free Monte Carlo baseline}

\emph{What can we get from labels alone?} 
The model-free Monte Carlo baseline is a batch estimator that follows the classical expansion estimator under simple random sampling \cite{cochran1977sampling}:
\begin{equation}
    \hat F_{1:t}^{\mathrm{MC}}=F(\mathcal W)+\frac{N-w}{t}\sum_{i=1}^{t}f(s_i).
\label{eq:monte_carlo}
\end{equation}
The second term scales the mean of the sampled units to estimate the total of the $N-w$ non-warmup units.

\subsubsection{Model-assisted uniform sampling estimators} 

\emph{Can uniform samples correct the model's total?} 
Let $M_t=|\mathcal R_t|$. We consider the incremental estimator family
\begin{equation}
    \hat F_t^{(c_t)} = F(\mathcal D_t) + c_tG_t(\mathcal R_t) + M_t[f(s_t) - c_tg_t(s_t)]
\label{eq:incremental_correction}
\end{equation}
parameterized by coefficient $c_t$, 
where both $g_t$ and $c_t$ are fixed before $s_t$ is drawn uniformly from $\mathcal R_t$. So they may depend on everything revealed so far, but not on $s_t$.
We instantiate three members of this family, the \emph{incremental} difference, ratio, and regression estimators, which follow classical estimators by the same names~\cite{sarndal2003model} but are modified to be exactly unbiased under coefficients $c_t$ and model sequences $g_t$ that adapt to revealed samples.
The difference estimator uses $c_t = 1$, the ratio estimator $c_t = \hat \gamma_t = {F(\mathcal D_t)}/{G_t(\mathcal D_t)}$, and the regression estimator
\[
c_t=\hat\beta_t=\frac{\sum_{s\in\mathcal D_t}\left(g_t(s)-\bar g_t\right)\left(f(s)-\bar f_t\right)}{\sum_{s\in\mathcal D_t}\left(g_t(s)-\bar g_t\right)^2},
\]
where $\bar g_t$ and $\bar f_t$  are means over $\mathcal D_t$.
To the best of our knowledge, these estimators are novel. 
The classical ratio and regression estimators fit their coefficients on the full sample, which makes them biased; computing $\hat \gamma_t$ and $\hat\beta_t$ from the previously revealed $\mathcal D_t$ and holding them fixed for the fresh draw removes that bias. 
The final term is then an unbiased estimate of the residual total on $\mathcal R_t$ at every step and under any sampling history, so all three estimators are unbiased for $F_{\text{true}}$ regardless of the accuracy of $g_t$ (Appendix~\ref{appendix:estimator_math}).
Degenerate or under-determined coefficient fits revert to $c_t = 1$.

\subsubsection{Model-assisted importance sampling estimators} 

\emph{Can the model choose which unit to label next?} 
Importance sampling (IS) uses model predictions to prioritize units expected to contribute more to the collection total. 
DISCount implements this idea with a fixed proposal (DIS; \citealt{perez2023discount}), while active measurement \cite{hamilton2025activemeasurement} extends to sampling without replacement and an adaptive proposal (known as \emph{adaptive importance sampling} or AIS; \citealt{mcbook}) via the incremental estimator 
\begin{equation}
    \hat F_t^{\text{IS}}=F(\mathcal D_t)+{f(s_t)}/{q_t(s_t)}.
\label{eq:is}
\end{equation}
Because $q_t$ is likewise fixed before the draw, the second term is an unbiased estimate of the remaining total whenever $q_t(s)>0$ for every $s \in \mathcal R_t$.
The estimator has minimum variance (of zero) when $q_t(s)\propto f(s)$, which motivates sampling in proportion to the model's predicted counts.
We evaluate four proposal strategies that share this estimator: AIS, DIS, spotAIS, and spotDIS. 
Define
\begin{equation}
q_t(s) = \frac{\max\{g_t(s), \epsilon\}}{\sum_{u\in\mathcal R_t}\max\{g_t(u),\epsilon\}}
\label{eq:proposal}
\end{equation}
where $\epsilon > 0$ is used to ensure $q_t(s) > 0$.
AIS is the case where $g_t$, and hence $q_t$, is time-varying (i.e., the model is retrained), and DIS is the case where $g_t = g_0$ for all $t$ (i.e., the model is not retrained).
AIS and DIS use $\epsilon=1$ following prior work. 
We also introduce two new ``spot-checking'' variants, spotAIS and spotDIS, that ensure positivity through a different mechanism. 
They allocate a fraction $\lambda$ of the sampling budget to ``spot check'' units predicted to be empty and spend the rest in proportion to the model's counts:
\begin{equation}
    q_t=\lambda\, q_t^{\mathrm{spot}}+(1-\lambda)\,q_t^{\mathrm{count}}
\label{eq:proposal_spot}
\end{equation}
where $q_t^{\mathrm{spot}}$ is uniform over predicted-empty units, and $q_t^{\mathrm{count}}$ is Eq.~\eqref{eq:proposal} with $\epsilon=0$ restricted to the predicted-populated units; we use $\lambda=0.05$.
The two components have disjoint support, so every unit draws its probability from exactly one term. 
A unit is predicted empty when $g_t(s)=0$ or a lightweight binary logistic regression classifier $h_t$ flags it as empty. 
Trained on frozen DINOv3 \cite{simeoni2025dinov3} embeddings of the units in $\mathcal D_t$, $h_t$ refits in milliseconds and adds a representation-based signal beyond the detector's count.
See Appendix~\ref{appendix:spotcheck_proposal} for more details.

\subsubsection{Combining per-step estimates} \label{sec:combining}
For incremental estimators, each main-loop draw yields an unbiased per-step estimate of $F_{\text{true}}$, whether from $\hat F_t^{(c_t)}$ or from $\hat F_t^{\text{IS}}$.
They are combined to form an overall estimate $\hat F_{1:t}=\sum_{i=1}^{t}\bar\alpha_{i,t}\hat F_i$ using weights $\bar{\alpha}_{i,t} > 0$ that sum to one for each $t$.
Because these weights are fixed in advance and never depend on the observed labels, $\hat F_{1:t}$ is unbiased as well (Appendix~\ref{appendix:estimator_math}).
We use the ``COMB'' weighting scheme~\cite{hamilton2025activemeasurement} when models are fine-tuned, which gives more weight to later estimates based on two different factors: an improving model and a smaller remaining pool. We alternatively use the ``LURE'' weighting scheme when models are not updated \cite{farquhar2021statistical}.
Details are in Appendix~\ref{appendix:weights}.
The model-only and Monte Carlo baselines are batch estimators and do not require combination.

Table~\ref{tab:method_roster} summarizes the counting strategies we evaluate, represented by each estimator--sampler--model triple.

\section{Benchmark Setup}

\begin{table}[t]
\centering
\small
\setlength{\tabcolsep}{4pt}
\begin{tabular}{lll}
\toprule
Estimator & Sampler & Model \\
\midrule
Monte Carlo & uniform &  --- \\
\{model-only, diff, ratio, reg\} & uniform & \{fine-tuned, frozen\} \\
\addlinespace[2pt]
AIS, spotAIS & $q_t$ & fine-tuned \\
DIS, spotDIS & $q_t$ & frozen \\
\bottomrule
\end{tabular}
\caption{Estimator--sampler--model configurations evaluated. Uniform-sampling estimators are run with both fine-tuned and frozen models, while the IS variants are model-specific, with AIS paired to fine-tuned detectors and DIS to frozen models. Here diff, ratio, and reg denote the incremental difference, ratio, and regression estimators, defined in \S\ref{sec:estimators}.}
\label{tab:method_roster}
\end{table}

\begin{table}[t]
\centering
\small
\setlength{\tabcolsep}{4.5pt}
\begin{tabular}{lllrrr}
\toprule
Tier (\%$\emptyset$) & Dataset & Class & \%$\emptyset$ & $\mu$ & $N$ \\
\midrule
\multirow{3}{*}{\shortstack{full\\($<5\%$)}}
  & Malaria & trophozoite    & 0.0  & 2.42  & 655 \\
  & Malaria & difficult      & 0.0  & 1.29  & 347 \\
  & Malaria & red blood cell & 2.1  & 49.19 & 1{,}688 \\
\midrule
\multirow{3}{*}{\shortstack{partial\\($5$--$60\%$)}}
  & Birds & bird         & 24.4 & 14.36 & 1{,}302 \\
  & DOTA  & tennis court & 37.5 & 4.63  & 689 \\
  & DOTA  & harbor       & 52.1 & 3.90  & 2{,}084 \\
\midrule
\multirow{3}{*}{\shortstack{sparse\\($60$--$90\%$)}}
  & DOTA & ship          & 78.7 & 2.10 & 5{,}061 \\
  & DOTA & plane         & 79.4 & 0.67 & 5{,}043 \\
  & DOTA & small vehicle & 80.8 & 1.31 & 5{,}032 \\
\midrule
\multirow{3}{*}{\shortstack{rare\\($\ge 90\%$)}}
  & Wildlife & giraffe      & 91.7 & 0.16 & 5{,}005 \\
  & DOTA     & helicopter   & 92.8 & 0.47 & 1{,}523 \\
  & DOTA     & storage tank & 98.1 & 0.11 & 5{,}176 \\
\bottomrule
\end{tabular}
\caption{The 12 tasks of CLC-12. \emph{\%$\emptyset$} is the empty fraction, \emph{$\mu$} the average object count per unit, and \emph{$N$} the total number of units. Full per-task statistics are in Appendix~\ref{appendix:classsel}.}
\label{tab:roster}
\end{table}

\subsection{Datasets and Task Construction}\label{sec:tasks}

We build CLC-12 from four publicly available, exhaustively annotated detection datasets spanning distinct visual domains. 
\textit{Malaria Cell Images (BBBC041v1)}~\cite{ds_malaria} contains microscopy images annotated with seven cell categories: six states of malaria infection and a \emph{difficult} category for cells that cannot be confidently typed.
\textit{Birds}~\cite{hamilton2025activemeasurement} consists of image tiles from two high-resolution ground-level photographs of Tree Swallow flocks. 
\textit{Kenyan Wildlife Aerial Survey}~\cite{ds_kenyan_wildlife} comprises aerial survey photographs annotated for elephants, giraffes, and zebras. 
\textit{DOTA v1.5}~\cite{ds_dota_1.5} contains aerial and satellite imagery with annotations for sixteen categories of urban objects, such as planes and storage tanks.
Each dataset-class pair defines a counting task. For each task, we collect images containing the target object from the dataset and convert them into a set of estimation units. \cref{fig:tile_montage} shows examples of each task.

To standardize apparent object scale, we estimate a task-specific unit size from 30 annotated objects sampled from roughly ten whole images, choosing the resolution so that the median object appears at about 80 pixels after resizing the tile to the Faster R-CNN's standard resize target, 800 pixels. Images whose longer side is at most 1.5$\times$ the tile size are kept whole to avoid thin edge tiles; larger images are split into non-overlapping square tiles, cropped at the right and bottom edges. These 30 samples are used only for tiling configuration and do not count toward the labeling budget or warmup. Objects are assigned to the tile containing the center of their bounding box, preventing double counting and ensuring that counts sum to $F_{\text{true}}$ (details in Appendix~\ref{appendix:tiling}).

We characterize each task by its \emph{empty fraction}, defined as the proportion of units containing no target objects, to study how object sparsity in the collection relates to different counting strategies.
Across the four datasets, 27 classes are grouped into four occupancy tiers based on this fraction: \emph{full} ($<5\%$), \emph{partial} ($5$--$60\%$), \emph{sparse} ($60$--$90\%$), and \emph{rare} ($\ge90\%$).
From each tier, we select the three classes with the largest $F_{\text{true}}$ while preserving diversity in object type, yielding twelve tasks (Table~\ref{tab:roster}).

\subsection{Evaluation Protocol and Metrics}\label{sec:trajectory_simulation}
We evaluate every task and counting strategy combination by the absolute fractional error $\mathrm{AFE}_t \;=\; {\bigl|\hat F_{1:t}-F_{\text{true}}\bigr|}/{F_{\text{true}}}$ averaged over $1{,}000$ trials and traced from $t=1$ to $t=N/2$. 
Reporting the whole curve exposes how fast each strategy converges and where strategies cross over.
Frozen-model trials are cheap: each unit's prediction is computed once and never changes, so a single cached set of predictions $g_0$ per model--task pair can be reused across all trials.
Fine-tuned models are different, since the model is retrained as labels arrive and its predictions must be recomputed along every trajectory. 
Running $1{,}000$ such trials for all fine-tuned configurations would have required roughly 82 GPU-years on our compute. 
We therefore run 5 genuine adaptive trajectories per task--strategy pair and cache the model's predictions at each retraining step (every 25 steps), covering all units in $\Omega$, not only those sampled.
Each of the 5 cached sequences is then replayed under $200$ different sampling seeds, holding the model's adaptation path fixed while sampling randomness varies, yielding $1{,}000$ trials in total.
\citet{hamilton2025activemeasurement} used a similar approach with one genuine trajectory.
Appendix~\ref{appendix:trajectory_simulation} provides more details.

\section{Results}\label{sec:results}

\begin{figure}[t]
\centering
\includegraphics[width=0.9\columnwidth]{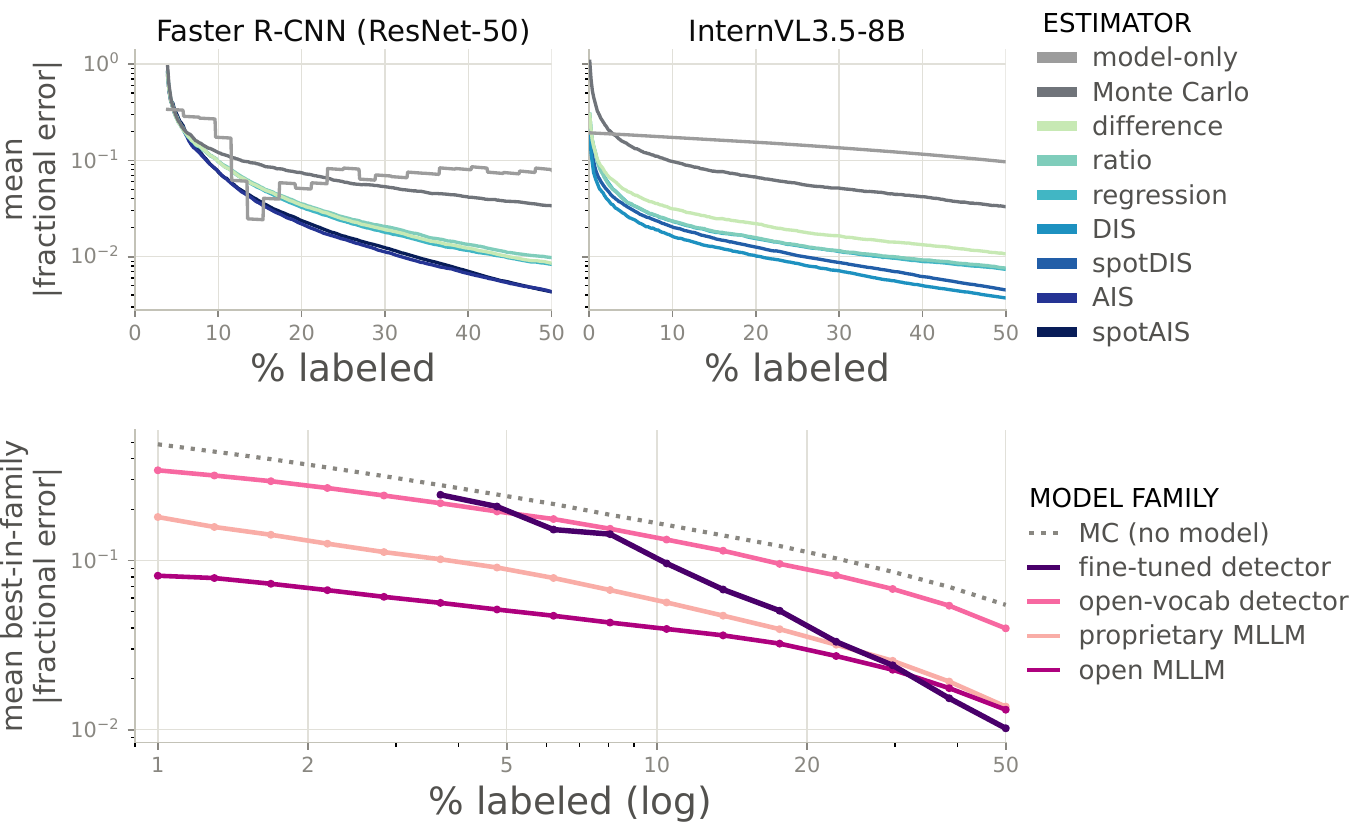}
\caption{
\textbf{Top:} mean absolute fractional error over 1{,}000 trajectories versus labeling budget on \textit{Birds/bird}, one curve per estimator, for fine-tuned Faster R-CNN (left) and frozen InternVL3.5-8B (right).
\textbf{Bottom:} each family's best case, averaged over 12 tasks; at every task and budget the family is represented by its lowest-error \{model, estimator\} pair, so a curve is a per-budget optimum, not one fixed model. The fine-tuned detector curve starts at $3.7\%$ labeled (warmup).}
\label{fig:fixed_and_crossover_birds_bird}
\end{figure}

\begin{figure*}[t]
\centering
\includegraphics[width=0.9\textwidth]{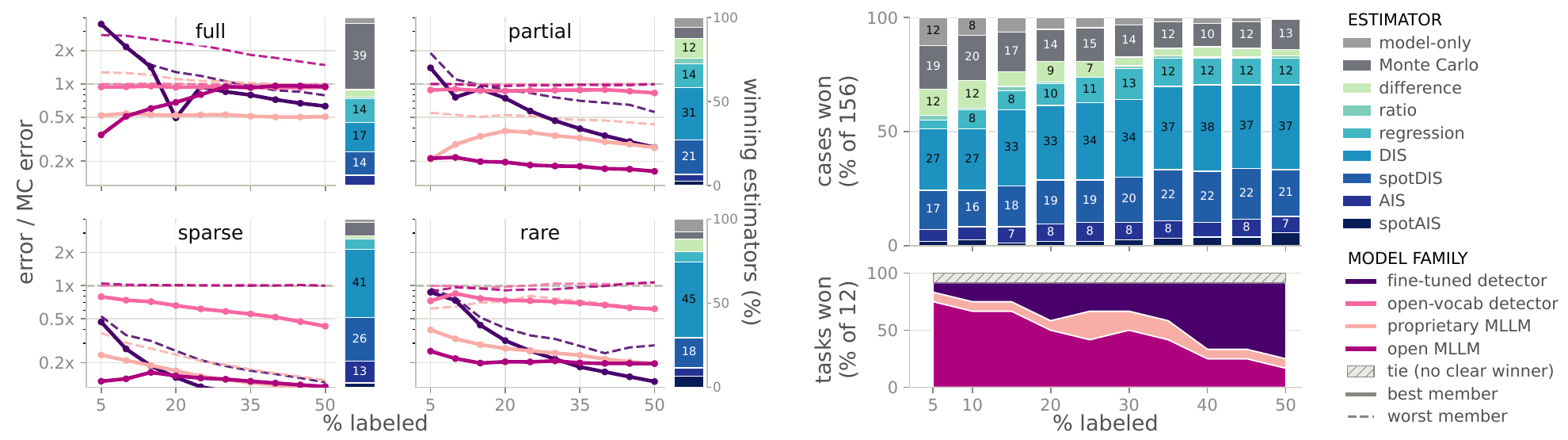}
\caption{Overview of the main results. A \emph{case} is one \{model, task, budget\} triple; its winner is the estimator (or, when comparing models, the model) with the lowest error on it.
\textbf{Left:} error relative to model-free Monte Carlo, one panel per occupancy tier (log scale; below $1\times$ the model helps). Every model is scored with its best estimator; the solid and dashed lines are the family's best and worst model, as medians over the tier's $3$ tasks. The bar beside each panel is the share (\%) of the tier's $390$ cases ($3$ tasks\,$\times$\,$13$ models\,$\times$\,$10$ budgets, $t=5\%,\ldots,50\%$) won by each estimator.
\textbf{Top right:} the same shares at each budget $t$, over the $156$ cases ($13$ models\,$\times$\,$12$ tasks) per $t$.
\textbf{Bottom right:} share of the $12$ tasks whose winning model belongs to each family, at each budget; the hatched band is the one task with no unique winner.}
\label{fig:results_overview}
\end{figure*}

\begin{figure}[t]
\centering
\includegraphics[width=0.9\columnwidth]{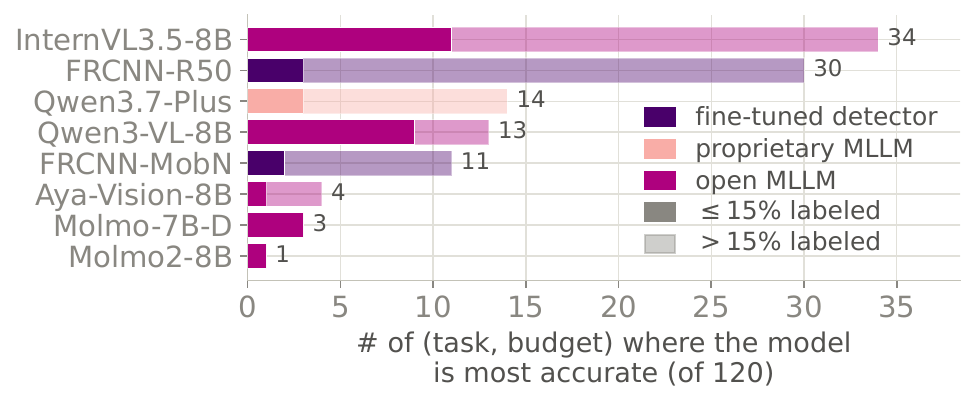}
\caption{Number of the $120$ (task, budget) settings on which each model, with its best estimator, has the lowest error of all $13$. The $10$ settings with no unique winner (see text) count for no model. Solid: won at $\leq15\%$ labeled; faded: above that. Models with no wins are omitted.
}
\label{fig:model_wins}
\end{figure}

The top row of \cref{fig:fixed_and_crossover_birds_bird} shows estimation error on the \textit{Birds/bird} task against the labeling budget. The ordering of curves previews what the rest of this section quantifies: importance sampling estimators reach the lowest error, uniform sampling estimators follow, and model-only is highest. For the frozen model, the model-only error declines only because revealed units replace predicted counts with true ones (\cref{eq:model_only}). Throughout, we put models on an equal footing by scoring each one with its best estimator at every task and budget, and each family by its best model.

\paragraph{Model-assisted counting is generally effective.} 
Model-assisted estimators outperform both the model-only and model-free Monte Carlo baselines on most tasks. We group models into four families: fine-tuned detector, OVD, proprietary MLLM, and open-source MLLM. The left panels of \cref{fig:results_overview} report each family's error relative to Monte Carlo across occupancy tiers (below one is better). The best model of every family beats Monte Carlo at nearly all budgets, and even the worst generally does, except on the full tier.

\paragraph{Open-source MLLMs are the best starter models; fine-tuned detectors overtake with more labels.}
The best model family changes with the labeling budget. In the bottom panel of \cref{fig:fixed_and_crossover_birds_bird}, the open-source MLLM curve is lowest from $1\%$ to about $30\%$ labeled. 
It is already under $10\%$ error at $1\%$ labeled, a budget at which Monte Carlo is still 3.8--6$\times$ worse. 
Fine-tuned detectors start high but improve most rapidly with labels, overtaking every frozen family by about $40\%$. 
The bottom-right panel of \cref{fig:results_overview} shows the same crossover task by task. A model wins a task outright when, scored with its best estimator, its error is strictly the lowest of the $13$ models. 
On \textit{Malaria/difficult} (hatched band), several models predict $0$ or $1$ on every unit, so the sampling floor gives them the same uniform proposal and identical estimates, and no model is credited (Appendix~\ref{appendix:classsel}).
At $5\%$ labeled, open-source MLLMs win $9$ of $11$ tasks; at $50\%$, detectors win $8$.
At the model level, \cref{fig:model_wins} shows that InternVL3.5-8B wins the most settings overall and at low budgets. Qwen3.7-Plus is the only proprietary MLLM that wins, and OVDs never win. Modern open-source MLLMs are thus a strong training-free starting point, and they substantially outperform earlier-generation OVDs, reflecting progress in vision models, though at a higher inference cost.

\paragraph{Importance sampling estimators are the best default.}
Labeling where the model predicts the most objects tends to beat correcting a uniform sample.
The estimator bars of \cref{fig:results_overview} show which estimator wins each case, by budget (top right) and by occupancy tier (beside each panel). 
Importance sampling estimators win $63\%$ of the $1{,}560$ cases: DIS wins most frequently at every budget, taking $40\%$ of the $1{,}320$ frozen models' cases, and AIS wins $46\%$ of the $240$ fine-tuned detectors' cases (the case imbalance makes AIS look minor in the pooled bars). Monte Carlo wins only $14.6\%$ of cases. 
The tier bars show that DIS's share rises from $17\%$ of the full tier to $45\%$ of the rare tier, uniform sampling estimators win most cases on the full tier but a quarter or fewer on the sparse and rare tiers, and $67\%$ of Monte Carlo's wins are concentrated in the full tier. Together, they suggest that the efficiency of uniform sampling relative to importance sampling declines as the target becomes rare. 
Still, uniform sampling may be desirable since it makes multi-class estimation~\citep{hamilton2026multitargetestimation} and data management more straightforward.
We therefore suggest DIS or AIS as the default, with uniform sampling as a fallback on dense targets.

\paragraph{Whether to adapt depends on the resources at hand.}
Off-the-shelf models are a low-barrier entry point through widely available APIs. But proprietary MLLMs can be expensive, roughly \$4.60--\$9.80 per 1,000 units for the costliest models in our study, and may be unsuitable when data privacy is a concern. 
Open-source MLLMs can be run locally, and smaller ones cost far less, at roughly \$0.25 per 1,000 units. Yet they remain more than 10 times more expensive than Faster R-CNN (\cref{tab:inference_cost}). 
A fine-tuned detector instead pays in labels and training time. One round of retraining the model and recomputing its predictions for all $s \in \Omega$ takes roughly $4$ minutes for MobileNet and $8$ for the ResNet backbone on average, hours of GPU time over a full run (with a retraining interval of $25$ steps), whereas frozen models are scored once. 
Since the detector's gain also varies across tasks, adaptation is worthwhile when labeling budgets are large, lightweight adaptation is available, or training can be amortized across related deployments, as in our case study (\S\ref{sec:casestudy}). 
Fully fine-tuning an MLLM with 7B or more parameters is compute-intensive. Lightweight adaptation and distillation into smaller detectors are promising future routes.

\paragraph{A small labeled pilot helps select a frozen model.}
Across the eight MLLMs, raw predicted totals vary substantially from $0.004\times$ to $55\times$ the true count (\cref{fig:mllm_disagreement}). Assuming DIS as the estimator choice, we propose a way to select a model before the main loop in three steps: (i) label a pilot set $\mathcal P$ of 50 uniformly sampled units; (ii) score each candidate by the Hellinger distance between the DIS proposal $q$ (Eq.~\eqref{eq:proposal}) and the normalized true counts $p\propto f$ on $\mathcal P$; and (iii) keep the closest model.
We compare distributions because DIS has zero variance when $q=p$ (\S\ref{sec:estimators}).
At $10\%$ and $30\%$ labeled, the model selected from a single pilot is the best candidate under DIS in about half of pilots ($0.48$--$0.59$), with a median error of $1.00\times$ the best candidate's.
Averaged over pilots, the score also orders all candidates well (Spearman $\rho$ of $0.76$--$0.85$).
Per-unit RMSE and MAE ($\rho$ $0.36$--$0.53$) and the gap between predicted and true totals ($\rho$ $0.17$--$0.34$) trail Hellinger in ranking quality. 
We conclude that how the proposal spreads sampling mass across units predicts downstream DIS performance better than the accuracy of the predicted magnitudes (Appendix~\ref{appendix:pilot}).

\begin{figure*}[t]
      \centering
      \includegraphics[width=0.9\textwidth]{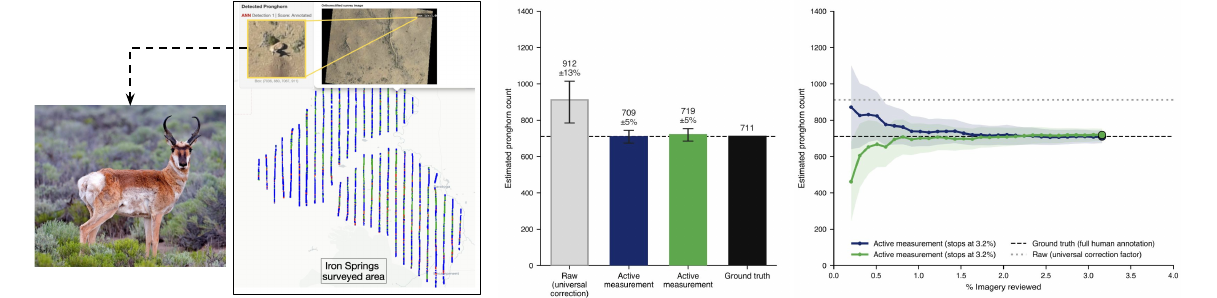}
      \caption{
      \textbf{Left:} a pronghorn (by Alan D. Wilson, CC BY-SA 3.0) and the Iron Springs survey area. Vertical lines are flight transects; each dot is the ground footprint of one survey image (blue: empty unit; green: empty but used for camera pose calibration; red: populated unit).
      \textbf{Center:} estimated pronghorn count from the calibrated model-only baseline (gray) and two AIS runs, with the ground truth. Whiskers are 95\% CI.
      \textbf{Right:} the two AIS trajectories and their 95\% CIs (shaded) versus \% of imagery reviewed.
      Both reached $\pm5\%$ uncertainty after reviewing 3.2\% of the imagery, the median over all 100 runs (\cref{fig:casestudy_grid}).
      }
    \label{fig:casestudy}
\end{figure*}

\section{Case Study: Pronghorn Aerial Surveys}\label{sec:casestudy}

Estimating how many animals are present in a given landscape is fundamental to robust wildlife management and conservation~\cite{williams2002analysis}. 
Aircraft make it possible to survey large, complex landscapes, but traditional aerial surveys, in which human observers fly at low altitude, are the most dangerous task that wildlife professionals perform~\cite{johnson1991use, sasse2003job}.
Image-based methods avoid this risk. Without observers on board, aircraft can fly faster and higher, and counts can be controlled for observer biases rather than depending on unrepeatable in-flight observation \cite{converse2024remote}.
Still, domain shift across landscapes and weather is unavoidable at scale \cite{converse2024remote}, so each survey’s population estimates must come with a quantified measure of reliability.

We deploy AIS on aerial imagery collected to estimate pronghorn (\textit{Antilocapra americana}) abundance across a 0.575 million-acre region of Wyoming, in collaboration with the Wyoming Game and Fish Department. Pronghorn inhabit open landscapes and are readily visible from above.
Where a traditional line-transect survey would have flown observers at $\sim$100\,m to sample 10\% of the landscape~\cite{johnson1991use}, the same aircraft flew a similar route at a safer 330\,m while a 45.7\,MP camera captured roughly one image per second, producing 6,129 images over about 44,000 acres. The collection falls in the \textit{rare} tier, with 98.8\% of images containing no pronghorn.

As in the benchmark, we use Faster R-CNN (ResNet-50 FPN, COCO-initialized). Unlike benchmark tasks which start from a small warmup set, a recurring survey can draw on data from earlier surveys. We therefore begin with a detector fine-tuned on four previous surveys from other regions of Wyoming (1,137 annotated tiles containing 1,864 pronghorn), which yields a better $g$ earlier in $t$.
We also use this data to build a better proposal. Each image and its detections are first split into eight non-overlapping tiles. Each tile is described by a fixed-length feature vector of the model's ranked confidence scores, whose length equals the maximum detection count across tiles.
A small regression model with a single hidden layer maps this feature vector to a predicted tile count. It is first trained on the earlier surveys and then retrained every 50 steps with the current data, which is far cheaper than fine-tuning the detector. Its predictions are then normalized to form $q_t$. This ranks predicted-empty tiles more finely than $g$ alone.
For comparison, we also evaluate a calibrated model-only baseline that multiplies $g$ by a precision-to-recall ratio estimated from the earlier surveys. We assess its stability by bootstrapping over those four surveys to obtain a distribution of calibration factors.

We ran AIS 100 times with different random seeds (see \cref{fig:casestudy_grid}) and show two representative runs in \cref{fig:casestudy} (labeled ``Active measurement''). Of the 100 runs, 96 reached a $\pm5\%$ confidence interval (CI) after reviewing only 3.0--3.4\% of the imagery. At the median of 3.2\%, this is about 1,570 eighth-image tiles, equivalent to roughly 200 full images or one afternoon of work. The other 4 runs stopped after reviewing 27--28.5\% of the imagery. Overall, 93 of the 100 CIs contain $F_{\mathrm{true}}$. Both example runs stop at the median and outperform the calibrated model-only baseline.
Partial landscape coverage contributes an additional $3\times$ uncertainty, setting a lower bound on the total survey uncertainty. This suggests that once image-level estimates are sufficiently reliable, reducing human effort may matter more than marginal reductions in estimate uncertainty. 
This case study shows how detector-development costs can be amortized across repeated surveys with lightweight adaptation addressing domain shift, and how each new survey can receive its own reliability estimate without exhaustive annotation.

\section{Discussion and Conclusion}
We systematically evaluated collection-level counting strategies across 12 tasks spanning biomedical microscopy, wildlife surveys, and overhead imagery of urban structures. Our results show that model-assisted estimation generally outperforms both model-only predictions and model-free Monte Carlo baselines, although the best strategy depends on the labeling budget and task structure. Importance sampling is a strong default when model predictions reliably rank high-count units. MLLMs offer an effective training-free starting point at low budgets, while fine-tuned detectors become increasingly competitive as more labels are acquired. Lightweight adaptation can further adapt models to deployment data without incurring the full cost of detector fine-tuning. Our findings suggest that models for collection-level counting should be selected not solely on raw counting accuracy, but also on how effectively their predictions support sampling and statistical estimation. Our pronghorn case study demonstrates that these gains translate to practice: AIS achieved a $\pm5\%$ confidence interval around the true total after reviewing only $\sim$3.2\% of the imagery.

Our work also has several limitations. Although CLC-12 spans diverse visual domains and occupancy tiers, it is limited to 12 single-class counting tasks. The findings may not fully generalize to multi-class estimation, alternative error metrics, noisy annotations, or settings in which target objects are difficult to verify consistently. Moreover, our comparisons do not capture all deployment costs, including image acquisition, model hosting, annotation interfaces, and the engineering required for repeated adaptation. Finally, our results reflect a snapshot of current frontier models and may need to be revisited as these models continue to improve.

\bibliography{aaai2027}
\clearpage
\appendix
\renewcommand{\thefigure}{A\arabic{figure}}
\renewcommand{\thetable}{A\arabic{table}}
\renewcommand{\theequation}{A\arabic{equation}}
\setcounter{figure}{0}
\setcounter{table}{0}
\setcounter{equation}{0}

\section{Bias and Variance of Estimators}\label{appendix:estimator_math}

This section shows that every estimator except the model-only baseline is unbiased.
The main text evaluates the estimates by their fractional error, but these estimators also yield confidence intervals. Here we write the variance estimators that construct them and report how often the resulting intervals cover $F_{\text{true}}$.
We use the notation of \S\ref{sec:setup}, and let $\mathcal H_t$ denote the history at step $t$: the revealed units, their labels, and the fitted $g_t$, $c_t$, $q_t$, and $h_t$.
Given $\mathcal H_t$, the only randomness at step $t$ is the draw $s_t\sim q_t$ from $\mathcal R_t$.

\subsubsection{Unbiasedness of the incremental estimators}
Every incremental estimator of \S\ref{sec:estimators} belongs to one family, written with the per-unit residual $d_t(s)=f(s)-c_tg_t(s)$ as
\begin{equation}
\hat F_t = F(\mathcal D_t) + c_t\,G_t(\mathcal R_t) + \frac{d_t(s_t)}{q_t(s_t)}.
\label{eq:general_perstep}
\end{equation}
The incremental difference, ratio, and regression estimators are the cases $c_t\in\{1,\hat\gamma_t,\hat\beta_t\}$ under uniform sampling, where $q_t\equiv1/M_t$ and the last term becomes $M_t\,d_t(s_t)$, recovering Eq.~\eqref{eq:incremental_correction}.
AIS, DIS, spotAIS, and spotDIS are the case $c_t=0$, recovering Eq.~\eqref{eq:is}.

Given $\mathcal H_t$, the first two terms of Eq.~\eqref{eq:general_perstep} are constants, and the last term is an importance-weighted estimate of the residual total on $\mathcal R_t$.
With $q_t(s)>0$ on $\mathcal R_t$,
\[
\begin{aligned}
&\mathbb E\bigl[\hat F_t\mid\mathcal H_t\bigr]\\
&\quad= F(\mathcal D_t) + c_tG_t(\mathcal R_t) + \sum_{s\in\mathcal R_t} q_t(s)\,\frac{d_t(s)}{q_t(s)}\\
&\quad= F(\mathcal D_t) + c_tG_t(\mathcal R_t) + F(\mathcal R_t) - c_tG_t(\mathcal R_t)\\
&\quad= F(\mathcal D_t) + F(\mathcal R_t) \;=\; F_{\text{true}}.
\end{aligned}
\]
The last term removes, in expectation, exactly the error of the scaled prediction $c_tG_t(\mathcal R_t)$, whatever $g_t$ and $c_t$ are.
This conditional expectation is the same for every history, and $\hat F_t$ is bounded because $\Omega$ is finite, $q_t>0$ on $\mathcal R_t$, and $c_t$ is finite. So the law of total expectation gives $\mathbb E[\hat F_t]=F_{\text{true}}$.
We only need that $q_t$ is positive on $\mathcal R_t$ and that $g_t$, $c_t$, and $q_t$ are determined by $\mathcal H_t$, so unbiasedness holds however the model and proposal adapt; the accuracy of $g_t$ affects only the variance.

The fallback to $c_t=1$ (\S\ref{sec:estimators}) applies while fewer than $10$ units have been revealed, and also when $G_t(\mathcal D_t)<1$ for the ratio coefficient or when $g_t$ is constant on $\mathcal D_t$ for the regression coefficient.
It depends only on $\mathcal H_t$, so it keeps $c_t$ finite without affecting unbiasedness.

The Monte Carlo baseline (Eq.~\eqref{eq:monte_carlo}) is not a member of this family.
It is the classical expansion estimator applied to the $N-w$ non-warmup units, where $w$ matches the model it is compared with ($w=50$ for fine-tuned detectors and $w=0$ for frozen models), and is unbiased under uniform sampling without replacement~\cite{cochran1977sampling}.

The COMB and LURE weights (Appendix~\ref{appendix:weights}) are positive, sum to one, and depend only on $N$, $w$, and the step index rather than on any observed label. Thus their weighted sum is also unbiased:
\[
\mathbb E\bigl[\hat F_{1:t}\bigr]
= \sum_{i=1}^{t}\bar\alpha_{i,t}\,\mathbb E\bigl[\hat F_i\bigr]
= F_{\text{true}}.
\]

\subsubsection{Variance and confidence intervals}
For all incremental estimators, we use the ``simple'' conditional variance estimator of \citet{hamilton2025activemeasurement},
\begin{equation}
\widehat{\mathrm{Var}}_{1:t}^{\,\mathrm{simp}}=\sum_{i=1}^{t}\bar\alpha_{i,t}^{\,2}\bigl(\hat F_i-\hat F_{1:t}\bigr)^{2}.
\label{eq:var_simp}
\end{equation}
They also define an alternative that estimates the conditional variance separately at each step, and report that the simple form behaves better where coverage is poor.
Confidence intervals are $\hat F_{1:t}\pm z_{1-\alpha/2}(\widehat{\mathrm{Var}}_{1:t}^{\,\mathrm{simp}})^{1/2}$, where $z_{1-\alpha/2}$ is the $1-\alpha/2$ quantile of the standard normal distribution.

The two batch estimators are handled separately.
The Monte Carlo baseline takes the classical without-replacement variance with a finite-population correction~\cite[Ch.~2]{cochran1977sampling}, applied to the non-warmup units because the warmup is counted exactly:
\[
\widehat{\mathrm{Var}}_{1:t}^{\,\mathrm{MC}}=\frac{K\,(K-t)}{t}\,\hat\sigma_t^{2},
\]
where $K=N-w$ and $\hat\sigma_t^{2}$ is the sample variance of the $t$ observed counts.
The model-only baseline has no variance estimator or confidence interval: given the model snapshot and the revealed set, it is deterministic and has no correction term from which a sampling error could be estimated.

\subsubsection{Confidence interval coverage}
\Cref{tab:ci_coverage} reports how often the nominal $95\%$ interval from a single trial contains $F_{\text{true}}$, and how wide it is.
Excluding the spot-checked variants, coverage is below nominal but improves with budget (\cref{tab:ci_coverage_budget}): it averages $0.89$ over the ten budgets and rises from $0.82$ at the $5\%$ budget to $0.92$ at $50\%$, consistent with \citet{hamilton2025activemeasurement}.
Among these estimators, AIS undercovers most at the smallest budgets ($0.62$ at $5\%$), but is in line with the other estimators from $20\%$ onward.

\subsubsection{Coverage of the spot-checked variants}
On the sparse and rare tiers, the spot-checked variants have the lowest coverage in \cref{tab:ci_coverage} ($0.57$--$0.79$).
Here, most units are empty and predicted to be empty, so $q_t^{\mathrm{spot}}$ spreads its $\lambda=5\%$ of the probability mass over many units (Appendix~\ref{appendix:spotcheck_proposal}).
A populated unit wrongly routed there therefore receives a very low sampling probability and hence a very large importance weight if drawn.
Most trials never draw such a unit and underestimate both the total and its variance; the few that do overestimate both.
The two effects cancel on average, so the mean estimate over trials remains close to $F_{\text{true}}$ (mean relative error $-0.001$), but a single trial usually reports an interval too narrow to reach $F_{\text{true}}$.
Consistent with this, each spot-checked variant reports a narrower interval than its non-spot counterpart on these tiers.
This explanation is specific to the sparse and rare tiers. It does not account for the below-nominal coverage of AIS and spotAIS on the full and partial tiers.

\subsubsection{Relevance to the case study}
\begin{figure*}[t]
      \includegraphics[width=\linewidth]{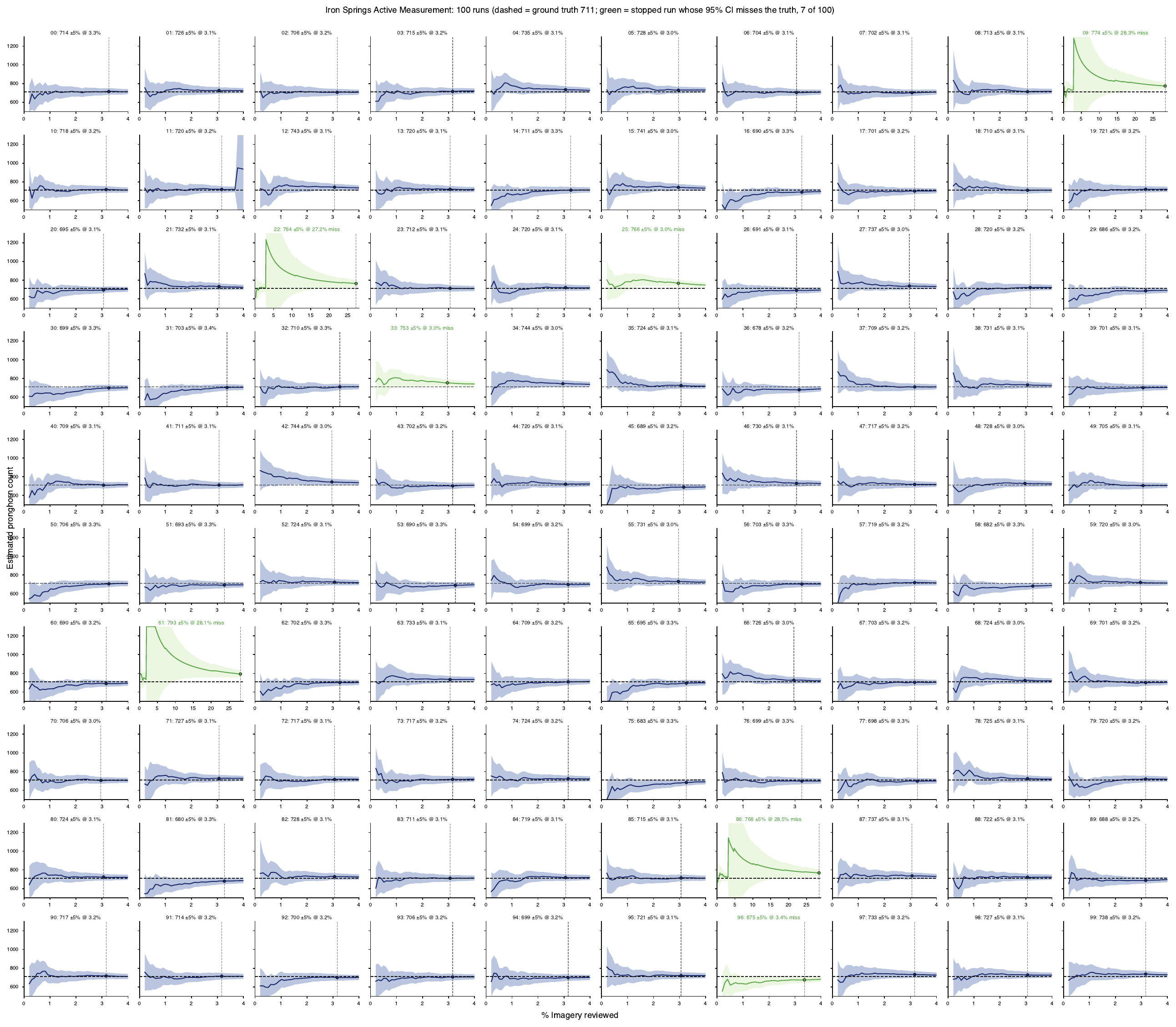}
      \caption{Trajectories of 100 independent AIS (active measurement) runs on the pronghorn survey, each with a different random seed. Every run reached the $\pm5\%$ estimated-uncertainty stopping condition: 96 after reviewing 3.0--3.4\% of the imagery (median 3.2\% over all 100 runs), and four after 27.2--28.5\%. Of the 100 final 95\% confidence intervals, 93 contain the true count; the seven that do not, including all four slow runs, are shown in green.}
    \label{fig:casestudy_grid}
\end{figure*}

These coverage results do not transfer to the pronghorn case study of \S\ref{sec:casestudy}.
It is a rare-tier collection, but its detector was fine-tuned on earlier surveys of the same species rather than starting from a small warmup, and its proposal is specifically adapted for the task to handle domain shift between deployments.
Its coverage is therefore measured separately: of $100$ AIS runs with different seeds, $93$ final $95\%$ intervals contain $F_{\text{true}}$ (\cref{fig:casestudy_grid}).
\begin{table}[t]
\centering
\small
\setlength{\tabcolsep}{2.5pt}
\begin{tabular}{lrrrrrrrr}
\toprule
& \multicolumn{4}{c}{\emph{Coverage}} & \multicolumn{4}{c}{\emph{Half-width}\,/\,$F_{\text{true}}$} \\
\cmidrule(lr){2-5}\cmidrule(lr){6-9}
Estimator & full & part & spar & rare & full & part & spar & rare \\
\midrule
Monte Carlo & 0.94 & 0.94 & 0.93 & 0.86 & 0.07 & 0.15 & 0.20 & 0.49 \\
difference  & 0.89 & 0.93 & 0.88 & 0.88 & 0.09 & 0.12 & 0.18 & 0.47 \\
ratio       & 0.87 & 0.92 & 0.88 & 0.86 & 0.12 & 0.12 & 0.16 & 0.43 \\
regression  & 0.90 & 0.92 & 0.88 & 0.86 & 0.07 & 0.11 & 0.16 & 0.41 \\
DIS         & 0.92 & 0.92 & 0.93 & 0.86 & 0.08 & 0.11 & 0.09 & 0.19 \\
AIS         & 0.83 & 0.86 & 0.92 & 0.87 & 0.06 & 0.07 & 0.04 & 0.12 \\
spotDIS     & 0.91 & 0.89 & 0.79 & 0.61 & 0.10 & 0.10 & 0.06 & 0.09 \\
spotAIS     & 0.79 & 0.80 & 0.71 & 0.57 & 0.08 & 0.07 & 0.03 & 0.04 \\
\midrule
excl.\ spot & 0.89 & 0.92 & 0.89 & 0.86 & 0.09 & 0.11 & 0.15 & 0.38 \\
\bottomrule
\end{tabular}
\caption{Behavior of $95\%$ confidence intervals by occupancy tier.
\emph{Coverage} is the fraction of the $1{,}000$ trials of a \{model, task, budget\} case (a \{task, budget\} case for Monte Carlo, which uses no model) whose interval contains $F_{\text{true}}$, averaged over the cases in a tier.
\emph{Half-width} is $1.96\,(\widehat{\mathrm{Var}}_{1:t}^{\,\mathrm{simp}})^{1/2}/F_{\text{true}}$ (with $\widehat{\mathrm{Var}}_{1:t}^{\,\mathrm{MC}}$ for Monte Carlo), summarized by its median within a case and again across cases.
The \emph{excl.\ spot} row pools all cases of the six estimators other than spotDIS and spotAIS.}
\label{tab:ci_coverage}
\end{table}

\begin{table}[t]
\centering
\small
\setlength{\tabcolsep}{2.5pt}
\begin{tabular}{lrrrrrrrr}
\toprule
& \multicolumn{4}{c}{\emph{Coverage}} & \multicolumn{4}{c}{\emph{Half-width}\,/\,$F_{\text{true}}$} \\
\cmidrule(lr){2-5}\cmidrule(lr){6-9}
Estimator & $5\%$ & $10\%$ & $20\%$ & $50\%$ & $5\%$ & $10\%$ & $20\%$ & $50\%$ \\
\midrule
Monte Carlo & 0.86 & 0.90 & 0.92 & 0.94 & 0.38 & 0.28 & 0.19 & 0.10 \\
difference  & 0.82 & 0.87 & 0.90 & 0.92 & 0.37 & 0.27 & 0.18 & 0.09 \\
ratio       & 0.81 & 0.86 & 0.89 & 0.91 & 0.41 & 0.31 & 0.19 & 0.10 \\
regression  & 0.81 & 0.86 & 0.89 & 0.92 & 0.36 & 0.26 & 0.17 & 0.08 \\
DIS         & 0.87 & 0.89 & 0.91 & 0.92 & 0.31 & 0.21 & 0.14 & 0.07 \\
AIS         & 0.62 & 0.80 & 0.90 & 0.92 & 0.22 & 0.17 & 0.12 & 0.03 \\
spotDIS     & 0.73 & 0.76 & 0.79 & 0.85 & 0.20 & 0.13 & 0.10 & 0.05 \\
spotAIS     & 0.47 & 0.63 & 0.74 & 0.80 & 0.12 & 0.08 & 0.05 & 0.02 \\
\midrule
excl.\ spot & 0.82 & 0.87 & 0.90 & 0.92 & 0.36 & 0.26 & 0.17 & 0.08 \\
\bottomrule
\end{tabular}
\caption{Behavior of $95\%$ confidence intervals by labeling budget, at four of the ten evaluated budgets ($5\%$ to $50\%$ in steps of $5\%$). All rows are defined and summarized over cases as in \cref{tab:ci_coverage}, but pooled across tiers.}
\label{tab:ci_coverage_budget}
\end{table}

\section{Spot-Checked Proposal Construction} \label{appendix:spotcheck_proposal}

The spot-checked proposals of \S\ref{sec:estimators} reserve a small uniform budget for units predicted to be empty.
This section describes how the classifier $h_t$ behind that prediction is fit and how its threshold is set.
The two variants differ only in $g_t$: spotDIS keeps $g_t=g_0$, while spotAIS updates $g_t$ as the detector is retrained.
In both, $h_t$ is refit on $\mathcal D_t$ every $25$ steps, with empty units ($f(s)=0$) as the positive class.
Until at least two empty and two populated units have been revealed, $h_t$ is left unfit and flags nothing, so only units with $g_t(s)=0$ are predicted empty.
When no remaining unit is predicted empty, all probability goes to $q_t^{\mathrm{count}}$; when none is predicted populated, all goes to $q_t^{\mathrm{spot}}$.
Hence, $q_t(s)>0$ for every remaining unit, the estimator is unbiased for any $h_t$, and errors in $h_t$ affect only the variance.

We set $h_t$'s threshold to limit that variance, exploiting an asymmetry between the two kinds of error.
A populated unit routed to $q_t^{\mathrm{spot}}$ receives a small probability and hence a large importance weight $f(s)/q_t(s)$ when drawn, whereas an empty unit routed to $q_t^{\mathrm{count}}$ costs little, since $f(s)=0$ contributes nothing to the estimate.
We therefore place the threshold just above the $(1-\phi)$ quantile of the scores of the revealed populated units, with $\phi=0.01$, so that at most a fraction $\phi$ of them would be flagged as empty.
Scores are out-of-fold, from stratified cross-validation over $\mathcal D_t$ with five folds (fewer when a class has fewer than five revealed units), because in-sample scores are over-confident.
While the number $n$ of revealed populated units is below $100$, $\lfloor\phi n\rfloor=0$ and none of them is flagged.

\section{Estimator Weight Sequences}\label{appendix:weights}

The combination weights of \S\ref{sec:combining} are $\bar\alpha_{i,t}=\alpha_i/\sum_{j=1}^{t}\alpha_j$.
When the model is fine-tuned, we use the COMB weights of \citet{hamilton2025activemeasurement}, the product of the LURE weight for without-replacement sampling~\cite{farquhar2021statistical} and the square-root rule for adaptive proposals~\cite{owen2020square}:
\[
\begin{aligned}
\alpha_i^{\mathrm{COMB}}&=\omega_i^{\mathrm{LURE}}\sqrt{i},\\
\omega_i^{\mathrm{LURE}}&=\frac{1}{M_i\,(M_i-1)},
\end{aligned}
\]
where $M_i=|\mathcal R_i|=N-w-i+1$. So $\omega_i^{\mathrm{LURE}}$ grows as the remaining pool shrinks.
When the model is not updated, we drop the square-root factor and use $\alpha_i=\omega_i^{\mathrm{LURE}}$.
The weights require $M_i\ge2$, which the budget cap $t\le N/2$ guarantees for every task.

\section{Tile Sizing and Pool Construction}\label{appendix:tiling}

We write \emph{image} for a source image before tiling; a \emph{unit} of $\Omega$ is either a tile or a whole image.

\subsubsection{Tile-size rule} This section describes how we choose each task's tile size.
Raw object size varies by nearly an order of magnitude across tasks: the median longer side of the sampled objects ranges from $24$\,px for \textit{Birds/bird} to $190$\,px for \textit{DOTA/tennis court}.
A single tile size would therefore present very different apparent object scales across tasks, confounding the comparison across occupancy tiers.
Let $m$ be the median longer side of the sampled boxes described below.
We use Faster R-CNN's internal resizing, which resizes each unit to a short side of $800$\,px (Appendix~\ref{appendix:frcnn_training}), as the reference because it is the only model type we fine-tune.
An object of size $m$ in a tile of side $\ell$ then appears at $800\,m/\ell$\,px, so the $80$\,px target gives
\begin{equation}\label{eq:tile}
  \ell \;=\; \mathrm{round}_{16}\!\left(\tfrac{800}{80}\,m\right)
        \;=\; \mathrm{round}_{16}(10\,m),
\end{equation}
where $\mathrm{round}_{16}$ rounds to the nearest multiple of $16$ to match detector strides.
We chose $80$\,px to stay well above $32$\,px, which is both the smallest anchor size of the detectors and the upper limit of COCO's ``small'' objects (area below $32^2$\,px$^2$).
An image whose longer side is at most $1.5\ell$ is kept whole, since tiling it would produce one full tile plus thin strips.
The size check is applied per image, so a task may contain both whole-image and tile units.
A larger image of width $W$ and height $H$ yields a $\lceil W/\ell\rceil\times\lceil H/\ell\rceil$ grid of tiles, which are square except at the right and bottom edges.

\subsubsection{Object samples for tile-size estimation}
For each task, images are visited in a seeded random order, and at most three randomly chosen bounding boxes are read per image until $30$ boxes from at least $10$ images are collected.
\textit{Birds} has only two images, so all of its boxes come from those two.
Only each box's longer side enters Eq.~\eqref{eq:tile}, so the sample reveals no unit count, and the sampled images stay in the pool.
As a diagnostic, we check that the apparent sizes $800\,x/\ell$ of the sampled longer sides $x$ have first and third quartiles between $32$ and $150$\,px.
Among the 12 tasks, only \textit{DOTA/plane} falls outside, with an upper quartile at $214$\,px; we keep the task and its tile size as is.

\subsubsection{Pool construction}
Tiling every available image would make some tasks too large to run the benchmark efficiently. For example \textit{DOTA/small vehicle} reaches $5{,}032$ units from only $64$ of its $1{,}179$ images.
We therefore cap each task at roughly $5{,}000$ units.
Images are admitted whole, in a seeded random order, until the running unit count first reaches the cap.
An image's unit count is computed from its dimensions alone, so admission does not depend on object counts.
$\Omega$ and $F_{\text{true}}$ are defined on the admitted images, fixed before any model is run, and shared by all methods.
The images sampled for tile sizing are taken from the front of the same order.

\section{Class Selection}\label{appendix:classsel}

Table~\ref{tab:candidates} lists every class in the four datasets; none was filtered out before tiering.
Each row is measured on that class's own pool $\Omega$ (Appendix~\ref{appendix:tiling}), so the empty fraction is a property of the (class, tile size) pair rather than of the class alone.
Because tile size normalizes apparent object scale, the tiers compare classes at roughly matched object size rather than matched tile area. Within each tier we rank by $F_{\text{true}}$, since a larger total makes the fractional error less sensitive to single objects: one object is $0.6\%$ of \textit{DOTA/container crane}'s total of $156$, but only $0.001\%$ of \textit{Malaria/red blood cell}'s $83{,}034$.
We take the three largest classes in the full, partial, and sparse tiers.
In the rare tier, the three largest are giraffe, helicopter, and elephant.
We skip elephant and the fourth-largest, zebra, and take storage tank instead, so that the tier does not contain two similar aerial wildlife tasks.
Storage tank also has the most dispersed count distribution in the benchmark ($\mathrm{CV}=14.2$).

\textit{Malaria/difficult} is the third-largest class in the full tier, and we keep it deliberately, as a task expected to be hard for any model.
Its class is defined by annotator uncertainty (cells that cannot be confidently typed), not by appearance, so it cannot be named in a text prompt, and several frozen models predict $0$ or $1$ on every unit (\S\ref{sec:results}).
Work that evaluates prompted models may therefore report on the other $11$ \emph{promptable} tasks of CLC-12.

\begin{table*}[t]
\centering
\footnotesize
\begin{tabular}{lllrrrrrrrr}
\toprule
Tier & Dataset & Class & \%$\emptyset$ & $\mu$ & CV & $N$ & $F_{\text{true}}$
     & img. pool & img. total & tile px \\
\midrule
\multirow{6}{*}{\shortstack{full\\\footnotesize($<5\%$ empty)}}
  & Malaria & \textbf{red blood cell} & 2.1  & 49.19 & 0.77 & 1{,}688 & 83{,}034 & 1{,}328 & 1{,}328 & 1{,}168 \\
  & Malaria & \textbf{trophozoite}    & 0.0  & 2.42  & 0.72 & 655     & 1{,}584  & 655     & 655     & 1{,}488 \\
  & Malaria & \textbf{difficult}      & 0.0  & 1.29  & 0.44 & 347     & 446      & 347     & 347     & 1{,}424 \\
  & Malaria & schizont                & 0.0  & 1.14  & 0.35 & 166     & 190      & 166     & 166     & 1{,}536 \\
  & Malaria & gametocyte              & 0.0  & 1.06  & 0.23 & 147     & 156      & 147     & 147     & 1{,}408 \\
  & Malaria & leukocyte               & 0.0  & 1.10  & 0.27 & 94      & 103      & 94      & 94      & 1{,}248 \\
\midrule
\multirow{6}{*}{\shortstack{partial\\\footnotesize($5$--$60\%$ empty)}}
  & Birds   & \textbf{bird}           & 24.4 & 14.36 & 1.48 & 1{,}302 & 18{,}696 & 2   & 2   & 240 \\
  & DOTA    & \textbf{harbor}         & 52.1 & 3.90  & 2.40 & 2{,}084 & 8{,}118  & 456 & 456 & 1{,}360 \\
  & DOTA    & \textbf{tennis court}   & 37.5 & 4.63  & 1.19 & 689     & 3{,}188  & 401 & 401 & 1{,}888 \\
  & Malaria & ring                    & 30.1 & 1.11  & 0.96 & 472     & 522      & 301 & 301 & 1{,}280 \\
  & DOTA    & soccer ball field       & 47.9 & 1.08  & 1.75 & 449     & 487      & 189 & 189 & 2{,}336 \\
  & DOTA    & ground track field      & 56.0 & 0.68  & 1.96 & 696     & 476      & 251 & 251 & 2{,}384 \\
\midrule
\multirow{9}{*}{\shortstack{sparse\\\footnotesize($60$--$90\%$ empty)}}
  & DOTA & \textbf{ship}           & 78.7 & 2.10 & 5.07 & 5{,}061 & 10{,}632 & 160 & 573     & 512 \\
  & DOTA & \textbf{small vehicle}  & 80.8 & 1.31 & 5.37 & 5{,}032 & 6{,}583  & 64  & 1{,}179 & 256 \\
  & DOTA & \textbf{plane}          & 79.4 & 0.67 & 3.21 & 5{,}043 & 3{,}370  & 90  & 269     & 544 \\
  & DOTA & large vehicle           & 85.4 & 0.54 & 4.28 & 5{,}006 & 2{,}682  & 73  & 703     & 304 \\
  & DOTA & swimming pool           & 88.7 & 0.25 & 4.07 & 5{,}022 & 1{,}274  & 131 & 288     & 432 \\
  & DOTA & bridge                  & 85.1 & 0.23 & 2.87 & 5{,}041 & 1{,}136  & 125 & 288     & 608 \\
  & DOTA & basketball court        & 78.6 & 0.64 & 2.76 & 1{,}046 & 672      & 155 & 155     & 1{,}152 \\
  & DOTA & baseball diamond        & 79.4 & 0.43 & 2.51 & 1{,}439 & 625      & 174 & 174     & 1{,}152 \\
  & DOTA & container crane         & 80.9 & 1.03 & 3.68 & 152     & 156      & 9   & 9       & 992 \\
\midrule
\multirow{6}{*}{\shortstack{rare\\\footnotesize($\ge 90\%$ empty)}}
  & Wildlife & \textbf{giraffe}       & 91.7 & 0.16 & 4.39  & 5{,}005 & 806 & 143 & 209 & 752 \\
  & DOTA     & \textbf{helicopter}    & 92.8 & 0.47 & 8.12  & 1{,}523 & 713 & 48  & 48  & 768 \\
  & Wildlife & elephant               & 94.9 & 0.12 & 6.11  & 5{,}000 & 608 & 97  & 195 & 624 \\
  & Wildlife & zebra                  & 95.9 & 0.12 & 7.19  & 5{,}112 & 595 & 42  & 189 & 384 \\
  & DOTA     & \textbf{storage tank}  & 98.1 & 0.11 & 14.24 & 5{,}176 & 549 & 26  & 262 & 272 \\
  & DOTA     & roundabout             & 97.3 & 0.03 & 6.15  & 5{,}064 & 145 & 58  & 249 & 384 \\
\bottomrule
\end{tabular}
\caption{All candidate classes, grouped by occupancy tier and ordered by
decreasing pool total $F_{\text{true}}$ within each tier. Selected benchmark
classes are shown in \textbf{bold}. $\mu$ is the average
target-object count per unit and \emph{CV} its coefficient of variation
across units (dispersion); $N$ is the total number of estimation units; $F_{\text{true}}$ is its exact target-object count within the $N$ units;
\emph{img. pool} and \emph{img. total} are the numbers of included and
available source images, respectively; and \emph{tile px} is the class-specific
tile side length. Pools may slightly exceed the $5{,}000$-unit target because
the final admitted image is retained in full.}
\label{tab:candidates}
\end{table*}

\section{Detector Training Details}\label{appendix:frcnn_training}

This section gives the details needed to reproduce the two fine-tuned Faster R-CNN detectors.

\subsubsection{Warmup set}
The warmup set $\mathcal W$ is drawn once per trajectory using ground-truth occupancy: $25$ units are sampled uniformly from the populated units, and the other $25$ uniformly from the rest of $\Omega$, without replacement.
This guarantees that at least half of $\mathcal W$ is populated.
Because the warmup counts toward the labeling budget, on tasks with few units it can exceed the smallest budgets (e.g., $14\%$ of the $347$ units of \textit{Malaria/difficult}), so detector results there are reported at the first step after the warmup; the detectors still do not win these tasks at those budgets, so the conclusions are unchanged.

\subsubsection{Implementation and inference}
Both detectors are the torchvision implementations with COCO\_V1 weights.
Each unit is preprocessed by the detector's default transform, which normalizes by ImageNet channel statistics and resizes the unit to a short side of $800$\,px with the long side capped at $1{,}333$\,px; we add no data augmentation.
We raise the cap on detections per unit from the torchvision default of $100$ to $300$, since five of the twelve tasks have units with more than $100$ objects (up to $219$).
All other inference defaults are unchanged, including non-maximum suppression at IoU $0.5$.
The predicted count $g_t(s)$ is the number of detections on unit $s$ with confidence at least $0.5$.

\subsubsection{Training targets}
A unit's annotations are filtered to the target class, so objects of other classes are unlabeled and act as negatives.
Each retraining step uses every unit revealed so far, including empty ones.
While counts follow the center-point rule (\S\ref{sec:tasks}), training targets are all boxes overlapping the unit, clipped to its bounds.
Under the center-point rule, an object straddling a boundary would be visible but unlabeled in the neighboring unit, teaching the detector to treat it as background.
The cost is that a straddling object can be detected in both units, so $g_t$ can overcount there.
This error enters the model-only baseline directly, but affects only the variance of the model-assisted estimators.

\subsubsection{Fine-tuning}
We fine-tune the feature pyramid, region-proposal network, RoI heads, and the deepest backbone stage (\texttt{layer4} for ResNet-50; the final four inverted-residual blocks for MobileNetV3-Large), and freeze all earlier backbone stages.
Each retraining step continues from the previous checkpoint, runs five epochs over the revealed units under the standard Faster R-CNN losses, and then recomputes the predictions for all of $\Omega$.
We use SGD with momentum $0.9$, weight decay $10^{-4}$, and batch size $4$, rebuilding the optimizer at each retraining step.
Learning rates are constant: $5\times10^{-4}$ for the region-proposal network and RoI heads, and $5\times10^{-5}$ for the feature pyramid and the unfrozen backbone stage.

\subsubsection{Computing infrastructure and cost}
Runs used one GPU per job (at least $16$\,GB of VRAM) on a heterogeneous cluster, under PyTorch~$2.7.1$ and torchvision~$0.22.1$ built against CUDA~$11.8$.
The benchmark comprises $360$ genuine detector trajectories: $2$ backbones $\times$ $12$ tasks $\times$ $3$ samplers (uniform, AIS, spotAIS) $\times$ $5$ seeds.
The seed fixes the warmup draw, the sampler, and the training shuffles; we do not enable deterministic GPU kernels, so a rerun reproduces the protocol but not bitwise-identical weights.
In job wall-clock time, the $360$ trajectories cost about $3{,}600$ GPU-hours ($804$ for MobileNetV3-Large and $2{,}806$ for ResNet-50), including dataset construction, the initial fit on $\mathcal W$, and the overhead of resuming preempted jobs.
Scaling each of the $72$ settings from $5$ to $1{,}000$ genuine trajectories at its measured mean cost gives the $82$ GPU-year estimate of \S\ref{sec:trajectory_simulation}.

\section{Trajectory Simulation Details}\label{appendix:trajectory_simulation}

Because genuine fine-tuned trajectories are expensive in time, most fine-tuned trials are replays (\S\ref{sec:trajectory_simulation}).
This section describes replay and tests whether it changes the results.

\subsubsection{Replay}
A genuine fine-tuned trajectory caches the predictions $g_t$ on all of $\Omega$ at every retraining step and, for spotAIS, the classifier $h_t$.
In a genuine trajectory, the revealed labels determine how the model retrains, which in turn affects later sampling.
A replay severs this feedback loop: it reuses the original warmup set and cached snapshots and varies only the sampling randomness.
A replay may therefore use snapshots trained on labels it did not itself reveal, in the proposal for AIS and spotAIS, in the correction terms for the model-assisted uniform sampling estimators, and directly in the model-only estimate.
Frozen-model trials are all genuine: $g_0$ is computed once and reused, and for spotDIS the classifier $h_t$ is cheap enough to refit within every trial.
The Monte Carlo baseline uses no model at all. It estimates the total by scaling up the average of the true counts it has seen, so its trials are also cheap to run genuinely.
It also provides an exact control for replay. Computed on a replay of a fine-tuned path, Monte Carlo ignores the cached snapshots, so the replay differs from a genuine trial only in its uniform draws, which follow the same distribution.
For each fine-tuned setting, one of the $200$ sampling seeds per cached model path reproduces the genuine trajectory and the other $199$ are replays, so the $1{,}000$ trials comprise five genuine trajectories and $995$ replays.

\subsubsection{Validation metrics}
We use two metrics, both computed from $\hat F_{1:t}$ at the $10$, $20$, $30$, $40$, and $50\%$ budgets.
Each has a known value under faithful replay, that is, when genuine trajectories and replays are exchangeable.

\emph{Held-out coverage} asks whether a genuine estimate looks like a replay of a different path.
For each genuine trajectory, we take the $4\times199=796$ replays of the other four paths of the same task and check whether the genuine estimate falls inside their central $90\%$ interval.
We report the fraction of the $60$ genuine trajectories (twelve tasks, five paths) for which it does, averaged over the five budgets.
Under faithful replay, this fraction should be about $0.90$.

\emph{Landing shift} asks whether a genuine estimate sits systematically above or below the replays of its own path, which removes between-path variation.
Let $X_r$ be the estimate of genuine trajectory $r$, and $\mu_r$ and $\sigma_r$ the mean and standard deviation of its $199$ replays.
The landing position $u_r=(X_r-\mu_r)/\sigma_r$ places the genuine estimate among its replays, in replay standard deviations.
Under faithful replay, $u_r$ has mean zero and a standard deviation close to one ($1.008$ if the estimates are normally distributed); we observed that the exact Monte Carlo control gives $0.97$.
The landing shift is the average of $u_r$ over the five budgets, then over the five trajectories of each task, and then over the twelve tasks, so it is zero in expectation under faithful replay.

Averaging makes the landing shift far less variable than a single $u_r$, so we judge it by its standard error rather than by the standard deviation of $u_r$.
Because trajectories from the same task may share a discrepancy, we compute the standard error from the twelve task means rather than the $60$ trajectories; it is approximately $0.14$.
Under faithful replay, the landing shift would then fall within two standard errors of zero, $\pm0.28$, roughly $95\%$ of the time, so we treat shifts in this range as consistent with faithful replay.

\begin{figure*}[t]
\centering
\includegraphics[width=\linewidth]{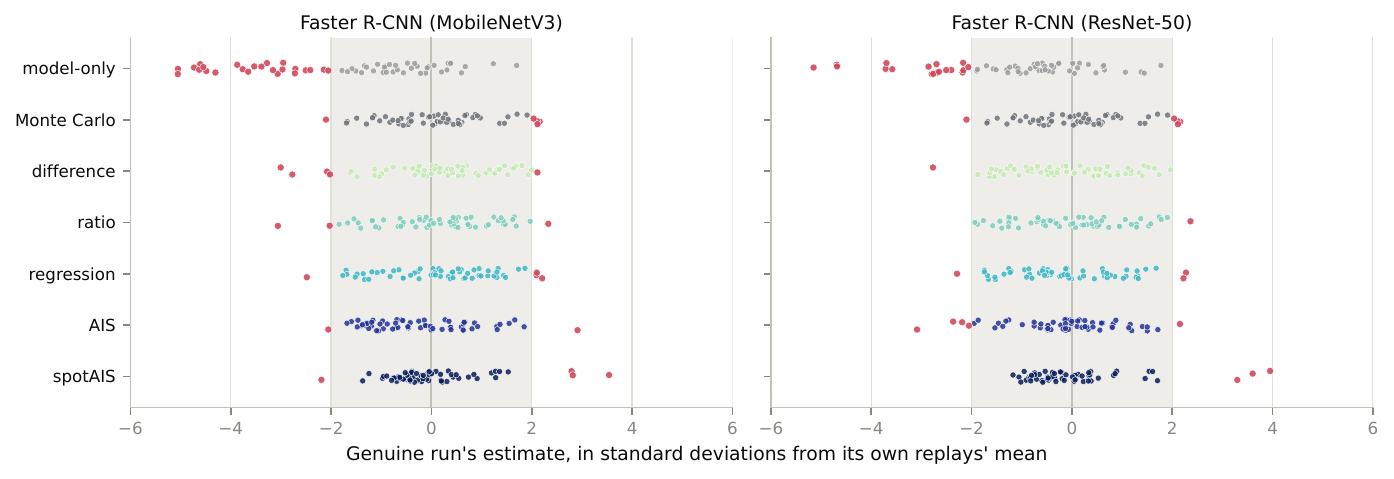}
\caption{Landing positions $u_r$ at the $30\%$ budget, underlying the \emph{Shift} column of Table~\ref{tab:replay_validation}. Each point is one genuine trajectory, placed relative to the $199$ replays of its own cached model path in replay standard deviations ($0$ is the replay mean). The shaded band spans $\pm2$ standard deviations for an individual trajectory, not the $\pm0.28$ interval for the mean shift; points outside it are red. The model-assisted estimators resemble the exact Monte Carlo control, whereas the model-only baseline has a pronounced negative tail.}
\label{fig:landing_positions}
\end{figure*}

\begin{table}[t]
\centering
\small
\setlength{\tabcolsep}{4pt}
\begin{tabular}{lcccc}
\toprule
 & \multicolumn{2}{c}{MobileNetV3} & \multicolumn{2}{c}{ResNet-50} \\
\cmidrule(lr){2-3}\cmidrule(lr){4-5}
Estimator & Cov. & Shift & Cov. & Shift \\
\midrule
\textit{faithful replay, in expectation} & \textit{.90} & \textit{0} & \textit{.90} & \textit{0} \\
\midrule
model-only                            & .61 & $-1.47$ & .64 & $-0.96$ \\
Monte Carlo \emph{(exact, measured)}  & .91 & $+0.10$ & .91 & $+0.10$ \\
difference                            & .86 & $+0.23$ & .89 & $+0.04$ \\
ratio                                 & .86 & $+0.17$ & .88 & $+0.10$ \\
regression                            & .88 & $+0.21$ & .88 & $+0.02$ \\
AIS                                   & .90 & $-0.01$ & .86 & $-0.06$ \\
spotAIS                               & .92 & $-0.03$ & .86 & $+0.05$ \\
\bottomrule
\end{tabular}
\caption{Replay validation for the two fine-tuned detectors. \emph{Cov.} is held-out coverage and \emph{Shift} is landing shift, both defined in Appendix~\ref{appendix:trajectory_simulation}. The first row gives their values under faithful replay; shifts within $\pm0.28$ are consistent with it. Monte Carlo ignores the detector, so its entries are the same under both.}
\label{tab:replay_validation}
\end{table}

\subsubsection{Fidelity for the model-assisted estimators}
Table~\ref{tab:replay_validation} reports both metrics, and \cref{fig:landing_positions} shows the landing positions at the $30\%$ budget.
For the model-assisted estimators, replay is close to genuine.
The Monte Carlo control gives a held-out coverage of $0.91$ and a shift of $+0.10$, illustrating the variation that remains with only five paths per task even when replay is exact.
Across the ten pairs of a model-assisted estimator (AIS, spotAIS, difference, ratio, or regression) and a detector, mean held-out coverage is $0.88$, and landing shifts range from $-0.06$ to $+0.23$, all inside the $\pm0.28$ range expected under faithful replay.
The largest shifts, $+0.23$ and $+0.21$ for difference and regression with MobileNetV3, are approximately $1.6$ and $1.5$ standard errors above zero.
These results do not establish that replay is equivalent to running $1{,}000$ genuine trajectories.
They indicate, however, that any discrepancy is small for these estimators: $0.03$ in mean held-out coverage relative to the exact control, and at most $0.23$ replay standard deviations in landing shift.

\subsubsection{Fidelity for the model-only baseline}
Replay is less faithful for the model-only baseline: its held-out coverage is only $0.62$ on average, and its shifts are $-1.47$ and $-0.96$ for MobileNetV3 and ResNet-50, the only table entries clearly outside sampling variation (see also the top row of \cref{fig:landing_positions}).
Genuine trajectories thus tend to produce lower estimates than their replays.
For MobileNetV3, the shift grows in magnitude from $-0.5$ at the $10\%$ budget to $-2.1$ at $50\%$, consistent with increasing divergence between a replay's revealed set and the set used to train its cached snapshot.
The model-only baseline has no correction term, so this mismatch enters the estimate directly.
At the $50\%$ budget, genuine trajectories have $9.4\%$ mean absolute fractional error, compared with $10.5\%$ for the replays, so our reported results may slightly overestimate the error of the model-only baseline.

\subsubsection{Limitations}
The analysis uses only five genuine trajectories per setting, which limits our estimate of between-path variability, especially at low budgets.
The coefficient of variation of the predicted pool total $G_t(\Omega)$ across the five paths exceeds $0.9$ on nine of the twelve tasks below the $20\%$ budget, and its median falls to $0.04$ from $30\%$ onward.
Replay estimates of the low-budget spread are therefore less certain than estimates of its center.

\section{Pilot Study Details}\label{appendix:pilot}

\begin{figure}[t]
\centering
\includegraphics[width=\linewidth]{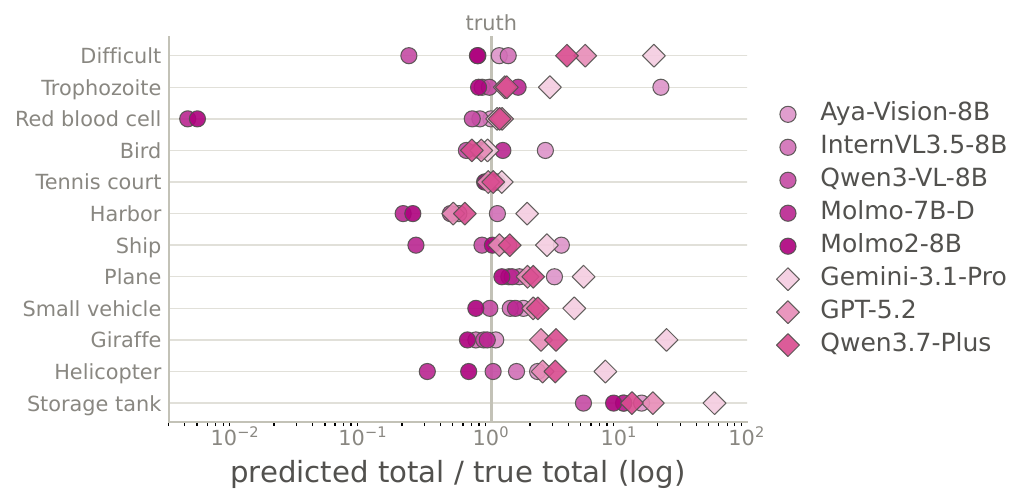}
\caption{Total count predicted by each of the eight MLLMs as a ratio to the true total, on every task (rows, ordered full to rare; log axis, $1 =$ exact); circles are open-source MLLMs and diamonds are proprietary ones, each tinted by model.}
\label{fig:mllm_disagreement}
\end{figure}

The eight MLLMs disagree widely: within a task their predicted totals can differ by orders of magnitude, and each model's accuracy shifts from task to task (\cref{fig:mllm_disagreement}).
Before the main loop, it can therefore be prudent to run a cheap labeled pilot that identifies the model likely to perform best on the task at hand.
This section describes the pilot procedure, the five proxies we compare, and how we evaluate them. The study covers only the $11$ frozen models; a one-shot pilot cannot represent a detector that retrains and improves as labels accumulate.
For each pilot, we draw $50$ units uniformly at random from $\Omega$, reveal their true counts, and score every candidate model on the same units.
Because a pilot is itself a random draw, we repeat the procedure over $10{,}000$ independent pilots per task, on two candidate sets (all $11$ frozen models, and the $8$ MLLMs alone) and at two labeling budgets ($10\%$ and $30\%$).

\subsubsection{The five proxies}
Let $f_i$ and $g_i$ be the true and predicted counts on pilot unit $i$.
Our default proxy is the Hellinger distance between the normalized true counts $p_i = f_i/\sum_j f_j$ and the proposal that DIS actually samples from, $q_i = \max\{g_i,1\}/\sum_j \max\{g_j,1\}$ (Eq.~\eqref{eq:proposal} with $\epsilon=1$),
\[
H(p,q) = \tfrac{1}{\sqrt{2}}\big(\sum_i (\sqrt{p_i}-\sqrt{q_i})^2\big)^{1/2}.
\]
We compare it with the same distance without the floor, $H(p,q^{\mathrm{raw}})$ with $q^{\mathrm{raw}}_i = g_i/\sum_j g_j$, which depends only on how the predicted mass is spread across units and not on its magnitude or the sampler. 
The other three proxies are scale-sensitive: the per-unit RMSE $\big(\tfrac{1}{|\mathcal{P}|}\sum_i (g_i-f_i)^2\big)^{1/2}$ and MAE $\tfrac{1}{|\mathcal{P}|}\sum_i |g_i-f_i|$, and the total-count gap $|\sum_i g_i - \sum_i f_i|/\sum_i f_i$.
The two Hellinger proxies and the total-count gap are undefined on a pilot that contains no objects ($H(p,q^{\mathrm{raw}})$ also when a model predicts zero on every pilot unit, which the floor in $q$ avoids). A model whose proxy is undefined is excluded from that pilot's selection, and $\rho$ averages each proxy over the pilots on which it is defined.

\subsubsection{Proxy evaluation}
Table~\ref{tab:pilot_proxies} evaluates each proxy with four metrics, all of which compare proxy values with the models' realized errors.
A model's realized error is the mean absolute fractional error of DIS with that model at that budget in our benchmark runs, because the pilot selects a model for DIS.
The first metric, $\rho$, measures how well a proxy orders all candidate models. We first average each model's proxy value over the $10{,}000$ pilots, and then take the within-task Spearman rank correlation between these averages and the realized errors.
Averaging before correlating removes pilot noise, so $\rho$ describes the proxy itself and is optimistic for a single pilot: computed pilot by pilot and then averaged, the default proxy's correlation is $0.62$--$0.70$ instead of $0.76$--$0.85$.
The other three metrics keep the pilot noise. Each is computed from the single model that one $50$-unit pilot picks, and only then summarized over pilots. \emph{beats} is the fraction of the other candidates whose error is higher than that of the pick (mean over pilots), \emph{win} is the fraction of pilots in which the pick has the lowest error among the candidates (a pick tied for the lowest counts as a win), and \emph{ratio} is the pick's error divided by that lowest error (median over pilots). For all four metrics, the ideal value is 1.0.

\subsubsection{Proxy comparison}
Among the five proxies, the default $H(p,q)$ is the best on every metric in every setting of Table~\ref{tab:pilot_proxies}, and the total-count gap is the worst. We take the metrics in order and compare the default with the scale-sensitive proxies.
\emph{Ordering.} The default has its clearest advantage in $\rho$, which is at least $0.76$ while no scale-sensitive proxy exceeds $0.53$. It orders all candidates rather than only identifying the best one, which matters when a practitioner wants to weigh several candidates.
\emph{Selection.} The default's pick has a \emph{beats} of at least $0.82$, against at most $0.76$ for RMSE and MAE.
Its \emph{win} is about one half, against at most $0.39$.
Its median \emph{ratio} is $1.00\times$ in every setting, against at least $1.11\times$. A \emph{ratio} of $1.00\times$ can accompany a \emph{win} below one half because it is rounded: with the $8$ MLLMs at $10\%$ labeled, the pick is the best model in $48\%$ of pilots and within $0.2\%$ of it in another $6\%$, so the median is $1.002\times$.
The median also hides the cost of a wrong pick, which is why we report \emph{win} alongside: the default's mean \emph{ratio} is $1.32$--$1.49\times$.
\emph{Effect of the floor.} Matching the proxy to the DIS proposal matters. Without the floor, $H(p,q^{\mathrm{raw}})$ is second among the five proxies on every metric and loses most in $\rho$, which is at most $0.66$.

\subsubsection{Alternative to Hellinger distance}
We also tried the chi-square divergence $\chi^2(p\,\|\,q) = \sum_i p_i^2/q_i - 1$, which on all of $\Omega$ equals the variance of the first DIS draw relative to $F_{\text{true}}^2$~\citep{sbert2022chisquare,muller2019neuralimportances}. This describes only the first draw. DIS samples without replacement, so each later draw is taken from the units that remain, and its variance follows the same formula with $p$, $q$, and the total computed over those units alone. The final estimate combines all per-step estimates, so its variance is a combination of these per-step variances and not the first-draw value itself; $\chi^2$ is therefore a proxy for the variance of DIS, not a prediction of it. With the same floored $q$, it is slightly more reliable than the default on our benchmark (last row of Table~\ref{tab:pilot_proxies}): its $\rho$ is higher by $0.04$--$0.05$, its \emph{beats} and \emph{win} are equal or slightly higher, and its \emph{ratio} is the same. The two pick the same model in $82\%$ of pilots on the MLLM + OVD set.
We keep the Hellinger distance as the default for two reasons, neither of which is selection performance. First, it is bounded and less sensitive to zeros in the proposal. It lies in $[0,1]$ whenever it is defined. Without the floor, it is undefined only when the model predicts zero on every pilot unit ($6\%$ of the pilot--model pairs in this study), whereas the chi-square divergence is infinite for $46\%$ of them, since a single unit that contains objects and receives zero proposal mass suffices. Second, its pilot estimate is more stable, with a median coefficient of variation across pilots of $0.12$ against $0.62$ for the chi-square divergence. With the floor $\epsilon=1$ used here, the chi-square divergence is a sound alternative.

\subsubsection{Conclusion}
For choosing a frozen model, we therefore recommend ranking candidates by $H(p,q)$ on the pilot. It rewards models whose proposal spreads sampling mass across units like the true counts, rather than the accuracy of their magnitude.

\subsubsection{Limitations} Three limits apply. First, we validate the proxy only for DIS with $\epsilon=1$; other floors and other proposals are outside this study.
Second, the floor makes the proxy depend on the scale of the predictions. A model that predicts counts in the hundreds is barely affected, whereas one that predicts $0$ or $1$ on most units is flattened toward a uniform proposal, so models with fractional or density-style outputs need $\epsilon$ chosen for their output scale.
Third, the pilot selects a model for DIS and does not select the estimator. At $10\%$ labeled, the DIS error of the pick is a median of $1.22$--$1.23\times$ the lowest error that any candidate reaches with any of the seven estimators available to a frozen model, and $1.01$--$1.02\times$ at $30\%$.

\subsubsection{Full-pool comparison}
\begin{figure*}[tp]
\centering
\includegraphics[width=\textwidth,height=0.33\textheight,keepaspectratio]{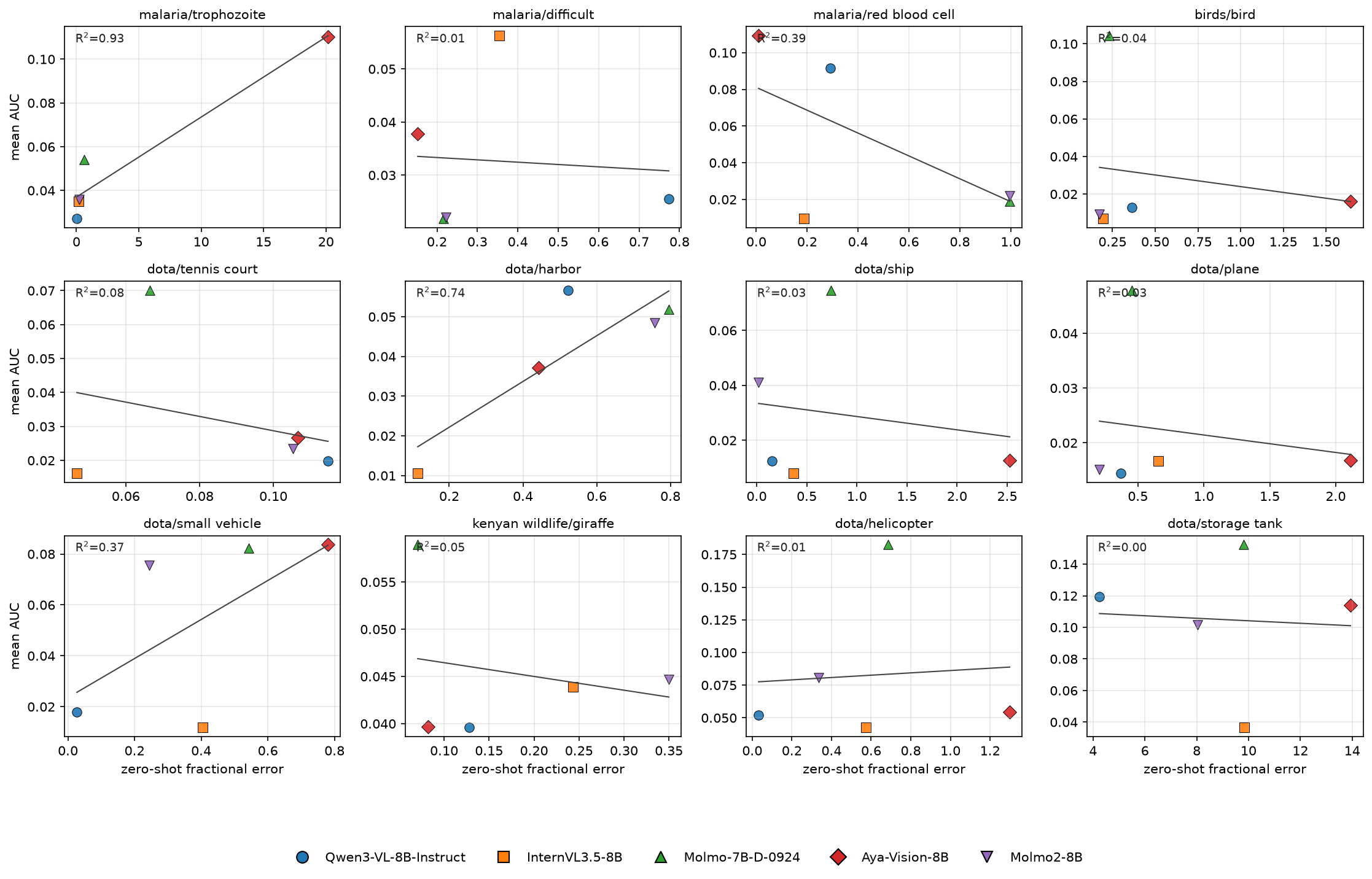}
\caption{Total-count gap on the full pool vs.\ mean AUC of DIS (area under the fractional-error curve up to $50\%$ labeled, averaged over $100$ trials). Each point is one frozen model on one task; lower is better on both axes.}
\label{fig:scatter_zeroshot_vs_auc}
\end{figure*}

\begin{figure*}[tp]
\centering
\includegraphics[width=\textwidth,height=0.33\textheight,keepaspectratio]{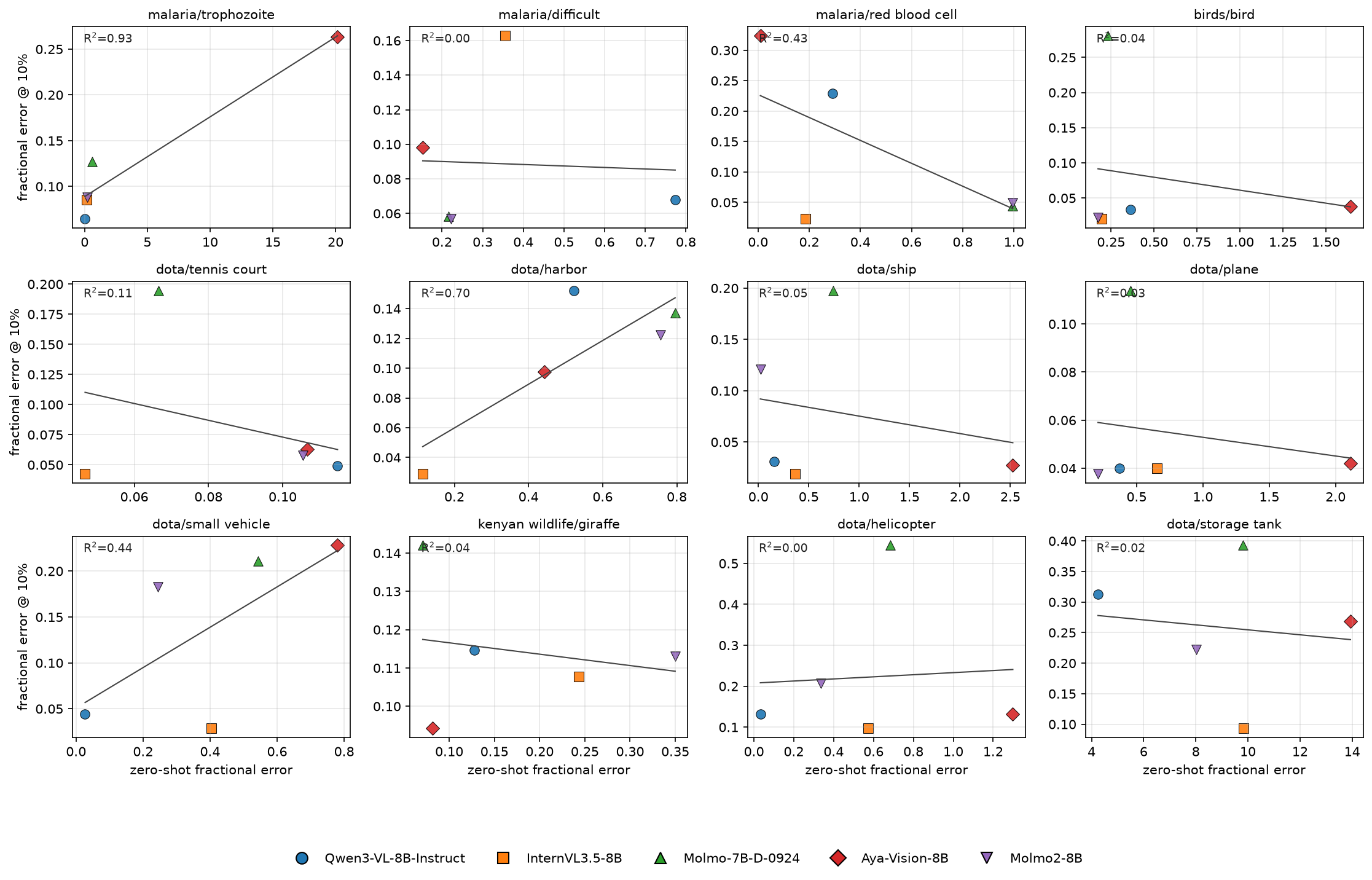}
\caption{Total-count gap on the full pool vs.\ fractional error of DIS at $10\%$ labeled. Each point is one frozen model on one task; lower is better on both axes.}
\label{fig:scatter_zeroshot_vs_err10}
\end{figure*}

\begin{figure*}[tp]
\centering
\includegraphics[width=\textwidth,height=0.33\textheight,keepaspectratio]{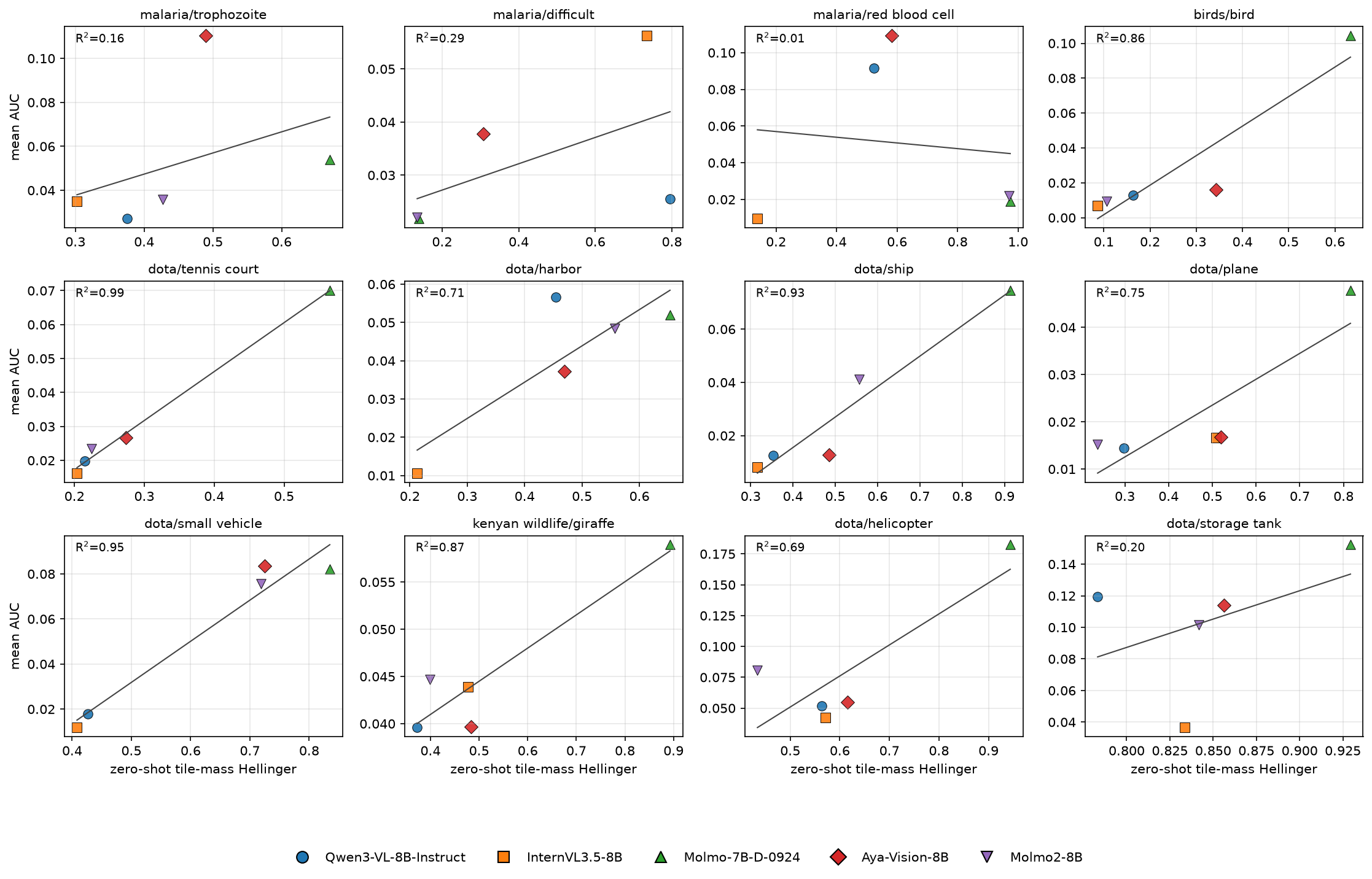}
\caption{Hellinger distance $H(p,q^{\mathrm{raw}})$ on the full pool vs.\ mean AUC of DIS (area under the fractional-error curve up to $50\%$ labeled, averaged over $100$ trials). Each point is one frozen model on one task; lower is better on both axes.}
\label{fig:scatter_hellinger_vs_auc}
\end{figure*}

\begin{figure*}[tp]
\centering
\includegraphics[width=\textwidth,height=0.33\textheight,keepaspectratio]{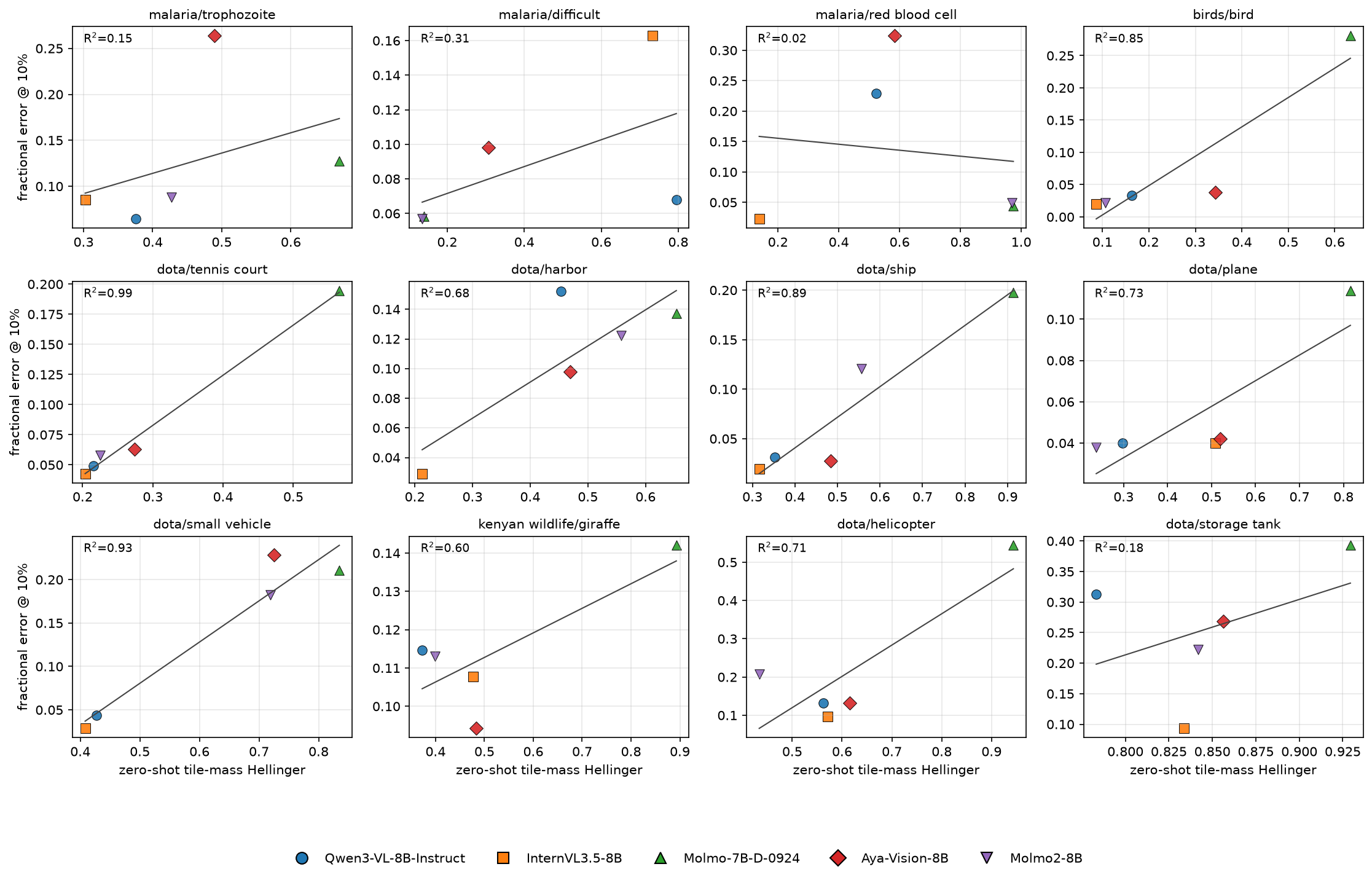}
\caption{Hellinger distance $H(p,q^{\mathrm{raw}})$ on the full pool vs.\ fractional error of DIS at $10\%$ labeled. Each point is one frozen model on one task; lower is better on both axes.}
\label{fig:scatter_hellinger_vs_err10}
\end{figure*}

\cref{fig:scatter_zeroshot_vs_auc,fig:scatter_zeroshot_vs_err10,fig:scatter_hellinger_vs_auc,fig:scatter_hellinger_vs_err10} compare the total-count gap and the Hellinger distance as predictors of downstream performance when both are computed on all of $\Omega$ rather than on a pilot, for each frozen model on each task. The Hellinger distance in these figures is the version without the floor, $H(p,q^{\mathrm{raw}})$.
Downstream performance is measured for DIS by the mean AUC, the area under the curve of fractional error against the percentage of units labeled (lower is better), and by the fractional error at $10\%$ labeled.
Against both measures, the Hellinger distance has a higher $R^2$ than the total-count gap $|G_0(\Omega) - F_{\text{true}}| / F_{\text{true}}$, so it explains more of the variation in downstream performance.
For example, on \textit{Birds/bird}, Molmo-7B-D-0924 has a smaller total-count gap than Aya Vision-8B but a much larger Hellinger distance, and a higher AUC and error at $10\%$ labeled.

\section{Model Inference, Cost, and Prompting} \label{appendix:mllm_prompting_details}

Here we compare the inference cost of all $13$ models, and describe how the MLLMs were run and how counts were extracted from their responses.
Table~\ref{tab:inference_cost} converts each model's cost to dollars per $1{,}000$ units, so that locally run and API-based models can be compared on one scale.
Table~\ref{tab:models_timing_cost} lists the raw measurements behind its MLLM rows (GPU-hours per unit or API cost per unit), along with each MLLM's source and count-extraction method.

\subsubsection{MLLM inference and cost estimation}

The open-source models (Qwen3-VL-8B, InternVL3.5-8B, Molmo-7B-D-0924, Aya Vision-8B, Molmo2-8B) were each run on a single A100 GPU over all units of the benchmark (Table~\ref{tab:models_timing_cost}).
Wall-clock time was measured from job start to the last prediction written, and totals approximately $15.4$ GPU-hours across the five models ($2.67 + 4.30 + 2.69 + 2.89 + 2.87$).
To measure latency and cost, the proprietary models (GPT-5.2, Gemini 3.1 Pro, Qwen3.7-Plus) were queried via the OpenRouter API on a sample of $60$ units ($5$ per task; $59$ for Qwen3.7-Plus due to one API failure), with cost computed from the recorded token usage and per-token pricing.
The samples cost \$0.2787 (GPT-5.2), \$0.5861 (Gemini 3.1 Pro), and \$0.3488 (Qwen3.7-Plus).
Extrapolated to the full benchmark, the estimated costs are \$156.58, \$329.35, and \$199.33, with estimated wall times of $60.2$, $67.3$, and $759.6$ hours.

\subsubsection{MLLM count extraction}
All five prompts are collected in Table~\ref{tab:prompts}, one per row.
For open-source models, the \emph{open-source count} prompt is followed by a second in-model pass (\emph{open-source count extraction}) that re-prompts the model to map its free-form response to a single integer, producing a structured choice (\texttt{nlg2choice}) parsed via \texttt{int(nlg2choice)} with a regex-digit fallback.
For proprietary models, the \emph{proprietary count} prompt asks the model to reason briefly and finish with a \texttt{Final answer: <number>} line. A separate extractor call (GPT-5.2 via OpenRouter) runs the \emph{proprietary count extraction} prompt, which uses the number on the final-answer line when present and falls back to a \texttt{final\_answer\_number} regex over the raw response. An optional \emph{verification} pass checks whether the extracted count is plainly present in the original response.

\begin{table*}[t]
\centering
\small
\setlength{\tabcolsep}{4pt}
\begin{tabular}{@{}l *{4}{cccc}@{}}
\toprule
 & \multicolumn{8}{c}{MLLM + OVD ($n=11$)} & \multicolumn{8}{c}{MLLM ($n=8$)} \\
\cmidrule(lr){2-9}\cmidrule(lr){10-17}
 & \multicolumn{4}{c}{$10\%$ labeled} & \multicolumn{4}{c}{$30\%$ labeled} & \multicolumn{4}{c}{$10\%$ labeled} & \multicolumn{4}{c}{$30\%$ labeled} \\
\cmidrule(lr){2-5}\cmidrule(lr){6-9}\cmidrule(lr){10-13}\cmidrule(lr){14-17}
Proxy & $\rho$ & beats & win & ratio & $\rho$ & beats & win & ratio & $\rho$ & beats & win & ratio & $\rho$ & beats & win & ratio \\
\midrule
$H(p,q)$ & \textbf{0.85} & \textbf{0.84} & \textbf{0.53} & \textbf{1.00} & \textbf{0.81} & \textbf{0.84} & \textbf{0.59} & \textbf{1.00} & \textbf{0.83} & \textbf{0.82} & \textbf{0.48} & \textbf{1.00} & \textbf{0.76} & \textbf{0.82} & \textbf{0.54} & \textbf{1.00} \\
$H(p,q^{\mathrm{raw}})$ & 0.66 & 0.79 & 0.47 & 1.02 & 0.62 & 0.78 & 0.46 & 1.06 & 0.65 & 0.77 & 0.42 & 1.02 & 0.56 & 0.75 & 0.41 & 1.06 \\
RMSE & 0.44 & 0.72 & 0.39 & 1.15 & 0.40 & 0.72 & 0.39 & 1.18 & 0.53 & 0.76 & 0.36 & 1.11 & 0.47 & 0.74 & 0.35 & 1.12 \\
MAE & 0.41 & 0.70 & 0.37 & 1.15 & 0.36 & 0.70 & 0.37 & 1.22 & 0.52 & 0.75 & 0.33 & 1.13 & 0.44 & 0.72 & 0.32 & 1.18 \\
Count gap & 0.21 & 0.53 & 0.13 & 1.74 & 0.17 & 0.53 & 0.12 & 1.89 & 0.34 & 0.58 & 0.12 & 1.59 & 0.26 & 0.55 & 0.11 & 1.58 \\
\midrule
$\chi^2(p\,\|\,q)$ & 0.89 & 0.85 & 0.58 & 1.00 & 0.86 & 0.85 & 0.59 & 1.00 & 0.87 & 0.84 & 0.53 & 1.00 & 0.81 & 0.84 & 0.54 & 1.00 \\
\bottomrule
\end{tabular}
\caption{Reliability of the five pilot proxies, and of the chi-square alternative (last row), for choosing a frozen model to run with DIS, on two candidate sets of $n$ models at two labeling budgets. The metrics are defined in Appendix~\ref{appendix:pilot}; the ideal value of each is $1$. $q$ is the DIS proposal, which floors each prediction at $1$, and $q^{\mathrm{raw}}$ is the normalized prediction without the floor. Bold marks the best of the five proxies in each column; the chi-square row is excluded from bolding.}
\label{tab:pilot_proxies}
\end{table*}

\begin{table*}[tp]
\centering
\setlength{\tabcolsep}{4pt}
\begin{tabular}[t]{lrrr}
\toprule
Model & \$/1{,}000 units & Units/\$1 & Normalized \\
\midrule
\multicolumn{4}{l}{\emph{Faster R-CNN}} \\
MobileNetV3 & \$0.017 & 59{,}507 & 581.3$\times$ \\
ResNet-50   & \$0.017 & 57{,}444 & 561.1$\times$ \\
\midrule
\multicolumn{4}{l}{\emph{Open-vocabulary detectors}} \\
Grounding DINO & \$0.126 & 7{,}913 & 77.3$\times$ \\
Florence-2     & \$0.410 & 2{,}440 & 23.8$\times$ \\
OWLv2          & \$0.578 & 1{,}729 & 16.9$\times$ \\
\bottomrule
\end{tabular}%
\hspace{1.5em}%
\begin{tabular}[t]{lrrr}
\toprule
Model & \$/1{,}000 units & Units/\$1 & Normalized \\
\midrule
\multicolumn{4}{l}{\emph{Open-source MLLMs}} \\
InternVL3.5-8B  & \$0.210 & 4{,}762 & 46.5$\times$ \\
Qwen3-VL-8B     & \$0.238 & 4{,}198 & 41.0$\times$ \\
Molmo-7B-D-0924 & \$0.240 & 4{,}167 & 40.7$\times$ \\
Molmo2-8B       & \$0.257 & 3{,}894 & 38.0$\times$ \\
Aya Vision-8B   & \$0.257 & 3{,}885 & 37.9$\times$ \\
\midrule
\multicolumn{4}{l}{\emph{Proprietary MLLMs}} \\
GPT-5.2        & \$4.645 & 215 & 2.1$\times$ \\
Qwen3.7-Plus   & \$5.912 & 169 & 1.7$\times$ \\
Gemini 3.1 Pro & \$9.768 & 102 & 1.0$\times$ \\
\bottomrule
\end{tabular}
\caption{
Per-unit inference cost, ordered from least to most expensive within each family.
\emph{Normalized} is units per dollar relative to Gemini 3.1 Pro. Locally run models assume one A100 at \$3 per GPU-hour;
proprietary MLLMs use the API cost per unit measured on the $60$-unit sample (Table~\ref{tab:models_timing_cost}).
}
\label{tab:inference_cost}
\end{table*}

\begin{table*}[ht!]
\centering
\begin{tabular}{lllccl}
\toprule
Model & Origin & Size & GPU-h/unit & API cost/unit & Extraction method \\
\midrule
Qwen3-VL-8B          & HuggingFace & 8B & $7.94{\times}10^{-5}$ & --- & Two-pass in-model count${\to}$choice \\
InternVL3.5-8B       & HuggingFace & 8B & $7.00{\times}10^{-5}$ & --- & Two-pass in-model count${\to}$choice \\
Molmo-7B-D-0924      & HuggingFace & 7B & $8.00{\times}10^{-5}$ & --- & Two-pass in-model count${\to}$choice \\
Aya Vision-8B        & HuggingFace & 8B & $8.58{\times}10^{-5}$ & --- & Two-pass in-model count${\to}$choice \\
Molmo2-8B            & HuggingFace & 8B & $8.56{\times}10^{-5}$ & --- & Two-pass in-model count${\to}$choice \\
\midrule
GPT-5.2              & OpenRouter  & --- & --- & \$0.004645 & ``Final answer:'' line + GPT-5.2 extractor pass \\
Gemini 3.1 Pro       & OpenRouter  & --- & --- & \$0.009768 & ``Final answer:'' line + GPT-5.2 extractor pass \\
Qwen3.7-Plus         & OpenRouter  & --- & --- & \$0.005912 & ``Final answer:'' line + GPT-5.2 extractor pass \\
\bottomrule
\end{tabular}
\caption{MLLM inference time and cost.
Open-source models ran on a single A100 over all $33{,}605$ units of the benchmark, and GPU-hours per unit are seconds per unit divided by $3{,}600$.
Proprietary models were queried via OpenRouter on a $60$-unit sample, and API cost per unit is the total API cost divided by the number of sampled units.}
\label{tab:models_timing_cost}
\end{table*}

\begin{table*}[t]
\centering
\small
\setlength{\tabcolsep}{4pt}
\begin{tabular}{@{}p{0.17\textwidth}p{0.32\textwidth}p{0.45\textwidth}@{}}
\toprule
\textbf{Prompt} &
\textbf{Placeholders and details} &
\textbf{Template} \\
\midrule

Open-source count
(stage 1)
&
The placeholder \texttt{\{object\_name\}} is replaced with the
task-specific object name (e.g.\ ``birds'', ``storage tanks'').
&
{\ttfamily
How many \{object\_name\} are in this image? Only output the number
and no other text.}
\\

Open-source count extraction
(stage 2, in-model second pass)
&
The placeholder \texttt{\{nlg\}} is the free-form text generated in
stage~1, and \texttt{\{max\_choice\}} is the upper bound of the
structured choice list (default 300). The vLLM batched variant omits
the ``from 0 to \{max\_choice\}'' clause.
&
{\ttfamily
What number was this response trying to say?\par
Response:\par
\{nlg\}\par
Answer only with one integer from 0 to \{max\_choice\}.}
\\

Proprietary count
(stage 1)
&
The placeholder \texttt{\{object\_name\}} is replaced with the
task-specific object name.
&
{\ttfamily
How many \{object\_name\} are in this image? Think briefly from
visible evidence, then finish with a line exactly like
Final answer: <number>.}
\\

Proprietary count extraction
(stage 2, GPT-5.2 extractor pass)
&
The placeholder \texttt{\{response\}} is the full reasoning text
produced in stage~1.
&
{\ttfamily
Extract the count from the response below.\par
Response:\par
\{response\}\par
Use the number on the final answer line when present. If the final
answer is a range or alternative like '2 or 3', output the first
number. Answer only with one number. If the final answer has no digit,
answer unknown.}
\\

Verification
(optional stage 3, GPT-5.2 verifier pass)
&
The placeholder \texttt{\{answer\_text\}} is the extracted count
(or ``unknown'') and \texttt{\{response\}} is the stage-1 reasoning
text.
&
{\ttfamily
Does the response explicitly contain this extracted count as its
answer?\par
Extracted count: \{answer\_text\}\par
Response:\par
\{response\}\par
Answer yes if the exact count is plainly present in the response.
Answer no if the response does not plainly contain that count.}
\\
\bottomrule
\end{tabular}
\caption{Prompts used for MLLM counting and response processing.
Open-source models use a count prompt followed by an in-model
count-extraction pass; proprietary models use a count prompt followed
by a GPT-5.2 extractor pass and, optionally, a GPT-5.2 verifier pass.}
\label{tab:prompts}
\end{table*}

\end{document}